%% file: 0_main.tex
\documentclass{article}

\usepackage[preprint]{neurips_2023}

\usepackage[utf8]{inputenc} 
\usepackage[T1]{fontenc}    
\usepackage{CJKutf8}
\usepackage{hyperref}       
\usepackage{url}            
\usepackage{booktabs}       
\usepackage{footnote}
\makesavenoteenv{figure}
\usepackage{amsfonts}       

\usepackage{nicefrac}       
\usepackage{microtype}      
\usepackage[table,xcdraw]{xcolor}         

\usepackage{caption}
\usepackage{wrapfig}
\usepackage{multirow}
\usepackage{makecell}
\usepackage{xspace}
\usepackage{enumitem}
\usepackage{amsmath}
\usepackage{listings}
\usepackage{algorithm}
\usepackage{algorithmic}
\usepackage{amsfonts,amssymb}
\usepackage{amsthm}
\usepackage{mathrsfs}
\usepackage{subcaption}
\usepackage{natbib}
\setcitestyle{numbers,square}
\usepackage{graphicx}       
\usepackage{cleveref}
\usepackage{fontawesome}
\usepackage{utfsym}
\usepackage[most]{tcolorbox}
\usepackage{siunitx} 
\usepackage{tabularx}
\usepackage{CJKutf8}
\usepackage{fancyvrb}
\usepackage{colortbl}
\usepackage{array}
\usepackage{threeparttable}
\usepackage{CJKutf8}
\usepackage{CJK}
\usepackage{booktabs}
\usepackage[table]{xcolor}
\usepackage{lipsum}
\usepackage{listings}
\definecolor{demphcolor}{RGB}{144,144,144}

\usepackage[most]{tcolorbox}

\newtcolorbox{lensbox}{
  colback=blue!5,
  colframe=blue!35,
  boxrule=0.5pt,
  arc=2pt,
  left=6pt,
  right=6pt,
  top=4pt,
  bottom=4pt,
  width=\linewidth,
  boxsep=0pt
}

\newcommand{\hhide}[1]{}
\newcommand{\hide}[1]{}

\newcommand{\model}{GLM-5V-Turbo}

\definecolor{zhipublue}{HTML}{3859FF}

\newtcolorbox{promptbox}[1][]{
  breakable,
  title=#1,
  colback=gray!5,
  colframe=black,
  colbacktitle=gray!15,
  coltitle=black,
  fonttitle=\bfseries,
  bottomrule=1.5pt,
  toprule=1.5pt,
  leftrule=1pt,
  rightrule=1pt,
  arc=0pt,
  outer arc=0pt,
  enhanced,
  before upper={\parindent=1.5em} 
}

\theoremstyle{definition}

\title{GLM-5V-Turbo: Toward a Native Foundation Model for Multimodal Agents}

 \author{
{GLM-5V-Turbo Team}
~\\\\
Z.ai ~\&~ Tsinghua University\\\\
(For the complete list of authors, please refer to the \hyperref[sec:contribution]{Contribution} section)
{}
}

\begin{document}

\maketitle

\vspace{-1.5em}

\begin{abstract}
We present GLM-5V-Turbo, a step toward native foundation models for multimodal agents. As foundation models are increasingly deployed in real environments, agentic capability depends not only on language reasoning, but also on the ability to perceive, interpret, and act over heterogeneous contexts such as images, videos, webpages, documents, GUIs. GLM-5V-Turbo is built around this objective: multimodal perception is integrated as a core component of reasoning, planning, tool use, and execution, rather than as an auxiliary interface to a language model.
This report summarizes the main improvements behind GLM-5V-Turbo across model design, multimodal training, reinforcement learning, toolchain expansion, and integration with agent frameworks. These developments lead to strong performance in multimodal coding, visual tool use, and framework-based agentic tasks, while preserving competitive text-only coding capability. More importantly, our development process offers practical insights for building multimodal agents, highlighting the central role of multimodal perception, hierarchical optimization, and reliable end-to-end verification.
\end{abstract}

\input{1_overview}

\input{2_model_training}

\input{3_multimodal_agentic}
\input{4_design_lenses}
\input{5_evaluation}
\input{6_remaining_challenges}

\clearpage

\input{7_contributors}

\clearpage

\bibliographystyle{abbrv}
\bibliography{ref}

\clearpage

\appendix

\input{8_appendix}

\end{document}

%% file: 1_overview.tex
\section{Overview}

Recent advances in foundation models have driven a shift from language understanding to agentic real-world interaction~\citep{anthropic2026claudeopus46,openai2026gpt54,zeng2026glm},
opening up substantial opportunities for productivity gains in domains such as knowledge work~\cite{google_gemini_deep_research_2025, openai_deep_research_2025, karpathy2026autoresearch}, software engineering~\cite{jimenez2023swe}, and tasks that require interacting with graphical user interfaces~\cite{hong2024cogagent,xie2024osworld}.
A general-purpose agentic model requires not only advanced intelligence, but also the ability to natively process complex multimodal context—including images, videos, text, webpages, and documents—and to integrate these heterogeneous inputs into a unified process of perception, reasoning, and decision-making~\cite{google_gemini_deep_research_2025,bytedance2026seed2,vteam2025glm45vglm41vthinkingversatilemultimodal}.

Toward this goal, we introduce a set of coordinated advances in model design, training, and infrastructure to enable more native multimodal modeling. In model design, we develop CogViT, a new vision encoder tailored for multimodal fine-grained understanding, and propose Multimodal Multi-Token Prediction, which supports both text-only and multimodal inputs while remaining friendly to large-scale infrastructure. 
In training, we deeply integrate vision and language throughout pre-training and supervised fine-tuning, and further perform joint reinforcement learning over more than 30 task categories spanning perception, reasoning, and agentic capabilities, supported by an optimized infrastructure stack for large-scale multimodal RL.
Building on these advances, we further expand \model{}'s multimodal agentic capabilities through toolchain extension, framework integration, and ecosystem development. We present a vision-centric deep search benchmark ImageMining that evaluates models' ability to ``think and deep search with image''.

These developments endow \model{} with native multimodal agentic capability, while retaining strong text-based agentic and coding performance relative to its language-only base model GLM-5-Turbo. 
This is reflected both in benchmark results and in its effectiveness in practical agentic settings, including chatbot-style environments such as Z.ai and framework-based scenarios such as Claude Code~\citep{anthropic2025claudecode} and OpenClaw~\citep{openclaw2026repo}.
\model{} achieves strong results on multimodal agentic benchmarks, including multimodal tool use (30.7 on ImageMining, 51.9 on BrowseComp-VL~\cite{BrowsecompVL}, 72.9 on MMSearch~\cite{MMSearch}, and 78.2 on SimpleVQA~\cite{SimpleVQA}), GUI agent tasks (75.7 on AndroidWorld~\cite{rawles2024androidworld} and 62.3 on OSWorld~\cite{xie2025osworld}), and Claw-based evaluations (87.0/80.7 on PinchBench~\cite{PinchBench}, 57.7/75.0 on ClawEval~\cite{ClawEval}, and 57.6 on ZClawBench~\cite{ZClawBench}). 
\model{} also demonstrates strong coding performance in both multimodal and text-only settings. For the multimodal setting, \model{} achieves 94.8 on Design2Code~\cite{si2025design2code}, outperforming Claude Opus 4.6~\citep{anthropic2026claudeopus46}; for the text-only setting, \model{} preserves the coding capability of its language-only base model GLM-5-Turbo and even surpasses it on CC-Backend (22.8), CC-Frontend (68.4), and CC-RepoExploration (72.2)~\cite{zeng2026glm}.

Developing \model{} also surfaced several broader lessons for agentic model development. Perception remains foundational to higher-level multimodal capability, while agentic competence is often acquired more effectively through hierarchical optimization than through monolithic end-to-end training. In addition, end-to-end agent tasks require clear specification, reliable verification, and carefully controlled evaluation for effective construction, assessment, and optimization. In this report, we summarize the main practices and lessons from developing \model{} to inform future work on native multimodal agents.

%% file: 2_model_training.tex
\section{Model, Training, and Infrastructure}
\subsection{CogViT Vision Encoder}
\begin{figure}[ht]
    \centering
    \includegraphics[width=0.9\textwidth]{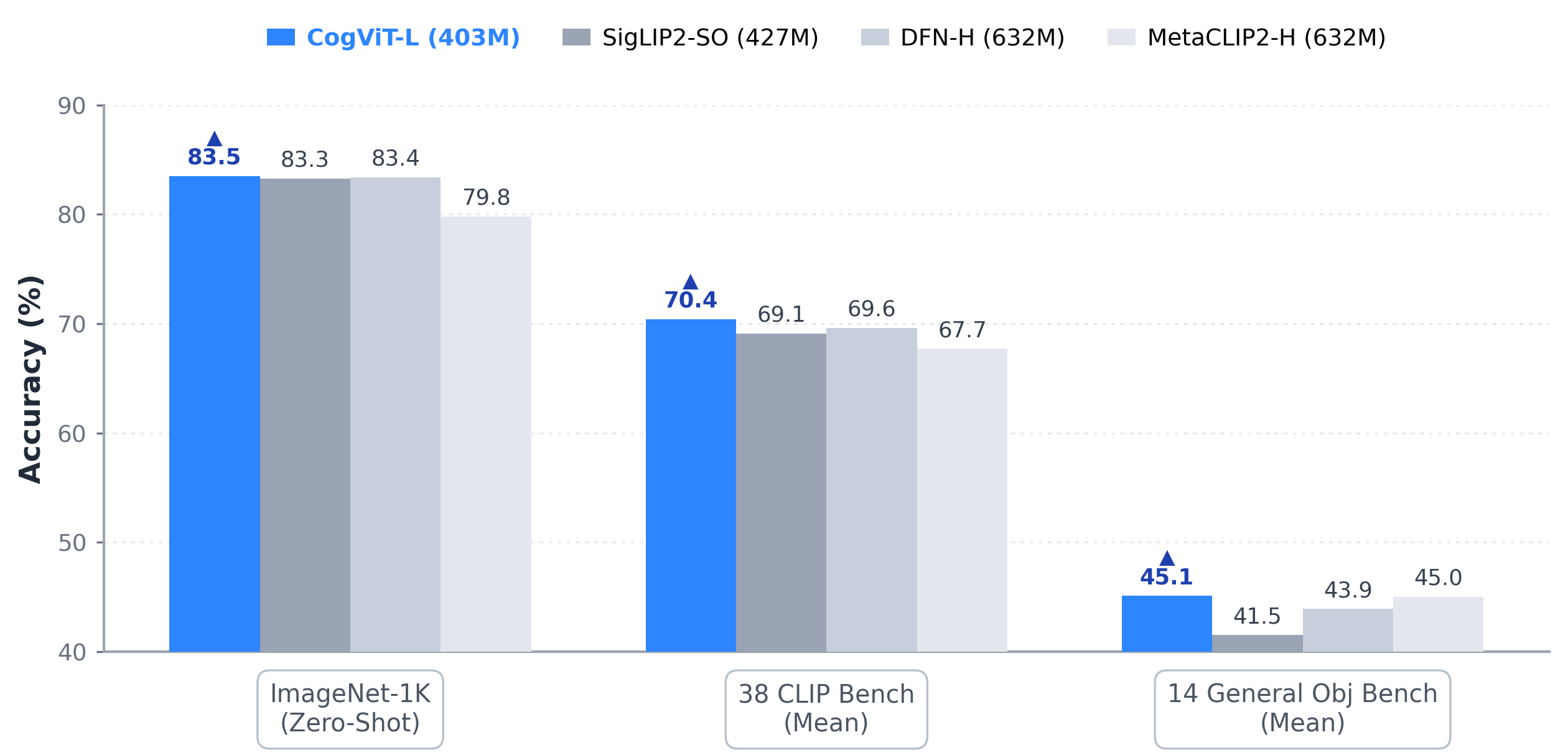} 
    \caption{Performance comparison of CogViT with other state-of-the-art vision encoders across general and fine-grained multimodal tasks.}
    \label{fig:cogvit_sota}
\end{figure}

We develop \textbf{CogViT}, a novel parameter-efficient vision encoder tailored for multimodal perception and downstream agent-oriented tasks. It delivers strong capabilities in general object recognition, fine-grained understanding, as well as geometric and spatial perception. As illustrated in Figure~\ref{fig:cogvit_sota}, CogViT achieves competitive performance across these domains. To balance representation learning with cross-modal alignment, we employ a two-stage pretraining recipe.

In the first stage, we use distillation-based masked image modeling to strengthen visual representations. Specifically, we train the student ViT to reconstruct the masked regions (35\% masking ratio, $224 \times 224$ resolution) in the feature spaces of dual teacher models: SigLIP2~\cite{tschannen2025siglip} for semantic representations and DINOv3~\cite{simeoni2025dinov3} for texture features. The training data follows a quality-aware mixture strategy: 80\% high-quality natural images, 10\% instruction-following data, and 10\% scientific imagery. We optimize with Muon~\cite{jordan2024muon} optimizer with a cosine decay schedule. Additionally, we introduce QK-Norm~\citep{henry2020query} to normalize query and key vectors before attention computation, effectively mitigating logit explosion and ensuring stability at scale.

The second stage shifts to contrastive image-text pretraining to align visual and textual features in a shared embedding space. Compared to the first stage, we introduce three key upgrades: (1) replacing the fixed $224 \times 224$ resolution with the NaFlex~\cite{tschannen2025siglip} scheme to process variable-size inputs while preserving original aspect ratios; (2) scaling the global batch size to 64K using the sigmoid-based SigLIP loss, combined with a bidirectional distributed implementation for efficiency; and (3) utilizing an 8-billion bilingual (Chinese-English) image-text corpus to enhance cross-lingual understanding. We continue to optimize with Muon, assigning module-specific learning rates and decay schedules to the vision, text, and projection components.

\subsection{Multimodal Multi-Token Prediction}
We propose \textbf{Multimodal Multi-Token Prediction (MMTP)}, a multimodal extension of multi-token prediction (MTP)~\citep{gloeckle2024mtp}, designed to support both text-only and multimodal inputs while remaining friendly to large-scale infrastructure.
The goal is to preserve acceptable length as well as training and inference efficiency in multimodal settings. 
In standard text-only MTP, prefix tokens can be passed into the MTP head directly through token IDs and embedded with the word embedding layer. 
Once MTP is extended to multimodal inputs, however, a central question arises: how should image tokens be passed to the MTP head? 
To answer this, we systematically compare three alternatives: The first directly passes the visual embeddings from the LLM backbone input to the MTP head; The second masks out all visual tokens at the MTP head input, reducing the design to text-only MTP; The third preserves visual positional information, but replaces all visual tokens with a shared learnable \texttt{<|image|>} special token as the visual input representation.

Considering both optimization behavior and system efficiency, GLM-5V-Turbo ultimately adopts the third design. 
Compared with directly passing visual embeddings to the MTP head, using the <|image|> token removes the need to propagate visual embeddings across pipeline-parallel stages, substantially reducing communication complexity while improving system scalability and engineering maintainability. 
Empirically, according to the ablation study on a 0.5B model, the \texttt{<|image|>}-based design achieves lower training loss and more stable convergence than directly using visual embeddings. We hypothesize that this is because the MTP head is typically lightweight, and may not have sufficient modeling capacity to effectively absorb visual representations whose distribution differs substantially from that of text embeddings; by contrast, the \texttt{<|image|>} token presents the input in a more uniform form and thus alleviates this optimization difficulty.
At the same time, compared with fully masking out visual tokens, this design remains naturally compatible with existing partitioning strategies such as sequence parallelism and context parallelism, without requiring additional handling for visual-embedding partitioning, alignment, or offset mapping, which reduces implementation complexity. Overall, the design gives GLM-5V-Turbo a more balanced trade-off among multimodal modeling capability, training stability, and system efficiency.

\begin{figure}[ht]
    \centering
    \includegraphics[width=0.8\textwidth]{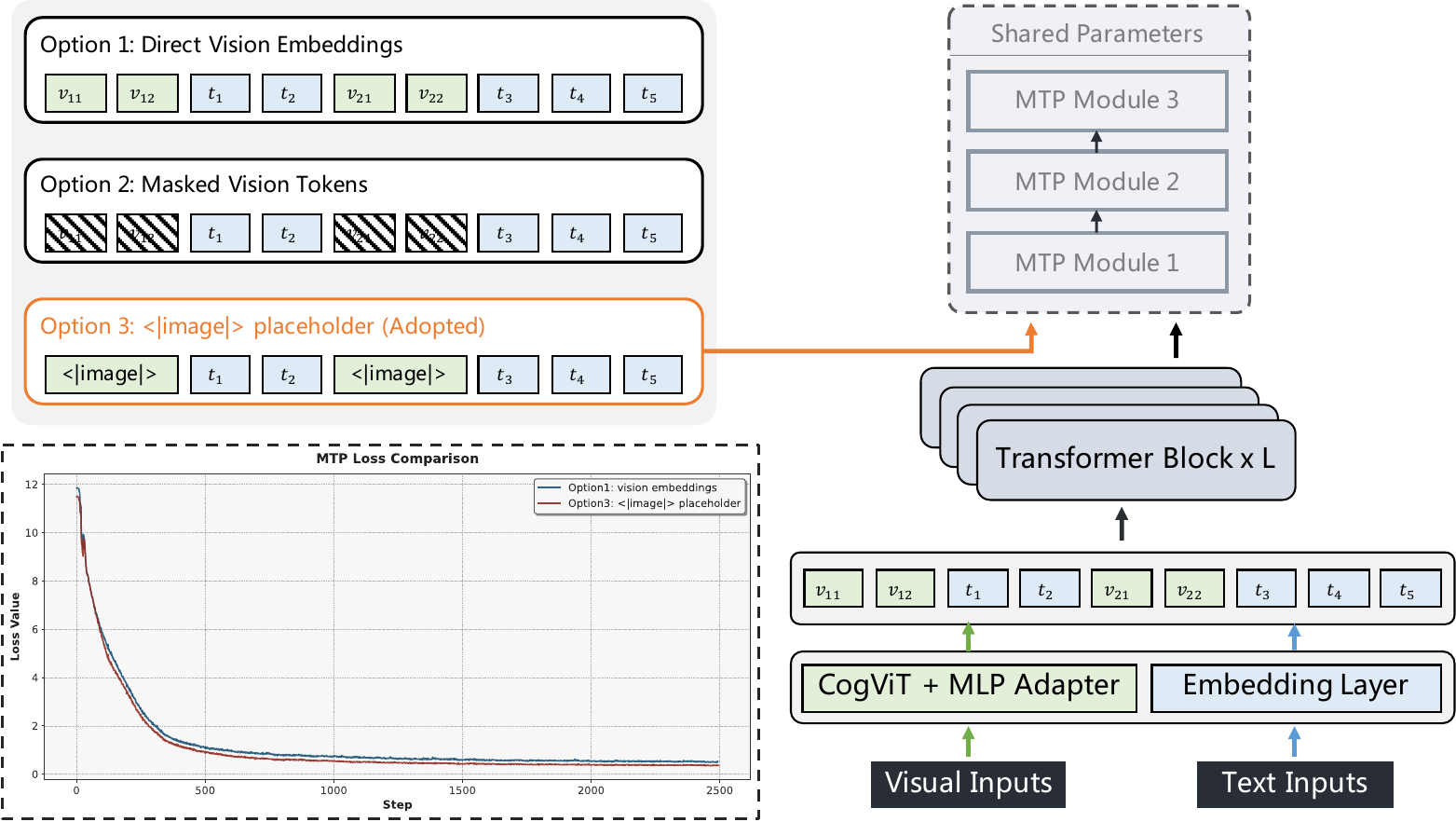} 
    \caption{Illustration of our multimodal multi-token prediction (MMTP) design. \textbf{Bottom-left:} Training loss curves comparing Option 1 and Option 3, where the adopted design achieves lower loss.}
    \label{fig:model_mtp}
\end{figure}

\subsection{Broad training across perception, reasoning, and agent capability}

The practical performance of multimodal agents depends on the joint development of perception, reasoning, planning, and execution, making narrow, domain-specific optimization insufficient. 
To improve these capabilities, we deeply integrate vision and language starting from the pretraining stage, strengthening the model’s native ability to represent and process multimodal context.
During the pre-training phase, we utilize a mixture of plain text and multimodal data to foster a balanced development of diverse capabilities. The multimodal datasets encompass a wide array of categories, including world knowledge, interleaved image-text, OCR, coding, GUI, video, multimodal tool-use, spatial perception, grounding, and academic problem-solving.
We place particular emphasis on multimodal coding data to better align visual understanding with code generation and to improve the model’s performance in multimodal agentic tasks.

GLM-5V-Turbo further undergoes \textbf{\textit{joint RL optimization over more than 30 task categories}}.
We adopt several technical improvements such as relative visual policy optimization in UI-to-code tasks~\citep{yang2025ui2code}.
This broad training setup yields gains at multiple levels: 
on the \textbf{perceptual} side, the model improves on tasks such as 2D image grounding and pointing (compared to SFT, the RL stage achieves improvements of 4.8\% and 3.2\% On RefCOCO-avg~\cite{kazemzadeh2014referitgame} and PointBench~\cite{cheng2025pointarena} respectively), video understanding (+5.6\% on MVBench~\cite{li2024mvbench}), 3D grounding (+7.7\% on SUNRGBD~\cite{song2015sun}), OCR (+4.2\% on OCRBench~\cite{liu2024ocrbench}), and chart understanding (+7.7\% on CharXiv~\cite{wang2024charxiv}); on \textbf{reasoning}-heavy tasks such as STEM (+1.8\% on MMMU\_Val~\cite{yue2024mmmu}, MMMU\_Pro~\cite{yue2025mmmu}, MathVista~\cite{lu2023mathvista} and LogicVista~\cite{xiao2024logicvista}), it exhibits greater stability in problem solving; and in \textbf{agentic} settings—including GUI agents (+4.9\% on OSWorld~\cite{xie2024osworld}), coding agents (+0.2\% on CC-Backend~\cite{zeng2026glm}), and general tool use (+3.5\% on MMSearch~\cite{jiang2024mmsearch} which demonstrates improved planning and execution).
Importantly, these gains are not confined to a single task family, but remain relatively consistent across a broad set of tasks.

This multi-task RL setting also exhibits several properties that we have consistently observed in earlier explorations such as GLM-4.1V-Thinking and GLM-4.5V~\citep{vteam2025glm45vglm41vthinkingversatilemultimodal}. 
Compared with the cross-domain trade-offs often seen in SFT, RL tends to show weaker interference across domains, allowing multiple domains to improve together with stable gains. 
Interestingly, in domains with narrower distributions where single-task RL is often prone to oscillation, collaborative training can make optimization more stable by exposing the model to a richer distribution of strategies and steering it toward more robust solutions.
Beyond this, we observe some transfer of thinking patterns across tasks: reasoning behaviors acquired in one domain can sometimes carry over to another and produce measurable benefits there as well. 
This suggests that the value of multi-task RL lies not only in covering a broader range of tasks, but also in inducing deeper sharing at the level of strategy patterns.

At the same time, broad coverage in joint optimization does not mean that the problem is fully resolved.
We do observe that capabilities left uncovered during RL can sometimes decline after post-training, especially those more orthogonal to the trained task distribution. 
One plausible explanation is that, as RL proceeds, both model capacity and learned thinking patterns become increasingly concentrated around the sampled task distribution, weakening the model’s ability to retain performance in under-represented domains. 
This suggests that the scope of task coverage during RL is itself an important factor shaping the model’s eventual generalization boundary. 
Even when a target capability cannot be easily formulated directly as an RL task, semantically or structurally related proxy tasks may provide useful optimization signals. 
For example, RL on single-turn UI-to-code generation can support more complex multi-turn coding ability. 
Taken together, these observations suggest that multi-task collaborative RL, including on-policy distillation, is not merely a tool for improving individual capabilities, but a central path toward shaping a more unified multimodal capability structure over a broader agentic distribution.

\subsection{Multimodal RL at Scale}

In the agent era, training infrastructure faces much stricter demands on both efficiency and stability, especially in large-scale multi-task multimodal reinforcement learning (RL).
Compared with conventional training, this setting must handle wide variation in prompt and response lengths, support both single-step and multi-step tasks, and coordinate one or more rule-based or model-based verifiers for each task. 
To address these challenges, we systematically redesign the training stack along four dimensions: unified task and reward abstraction, end-to-end asynchrony and stage overlap, fine-grained memory management for multimodal workloads, and topology-aware partitioning and load balancing for visual inputs.

\paragraph{Unified task and reward abstraction.}
We build a unified VLM RL Gym that provides a consistent environment interface for both single-step and multi-step tasks, so that heterogeneous task types can be handled within the same training framework. 
In parallel, we introduce an independent reward system that centrally orchestrates multiple verifiers. Rule-based verifiers are executed locally and synchronously, while model-based judges are invoked asynchronously through APIs; their outputs are then combined into rewards through configurable aggregation strategies, without entangling verifier logic with the main training codepath. To improve observability in mixed-task training, each sample also carries a data-source tag, allowing source-specific metrics such as reward and pass@k to be aggregated across parallel groups and reported separately.

\paragraph{Full-pipeline decoupling, asynchrony, and stage overlap.}
We restructure the training pipeline to decouple rollout inference, reward evaluation, batch construction and weight transfer, to maximize overlap across these stages. 
Each inference request is registered with a completion callback, so reward computation can be triggered as soon as that request finishes, rather than waiting for the entire rollout batch to complete; this reduces pipeline idle time caused by long-tail requests. 
Batch construction is executed in parallel with CPU–GPU transfer of old-policy weights. 
For the reference model, parameters remain resident on CPU memory, are asynchronously prefetched to GPU immediately before reference forward, and are released right after use, allowing reference computation to overlap effectively with the main training step. 
The system also supports two early-abort modes, based on either completion count or time threshold. Aborted prompts can be cached and reused, which helps control long-tail latency without materially reducing data utilization.

\paragraph{Fine-grained runtime memory management for multimodal workloads.}
Standard recomputation schemes are largely designed around text-only training and do not adequately address the memory bottlenecks introduced by multimodal inputs. 
To address this, we design separate memory-management strategies for the vision-side ViT and projector modules, combining targeted recomputation with CPU offloading. 
This prevents activation memory from scaling linearly with the number of images in the naïve way, and substantially reduces runtime memory pressure while preserving overall computational efficiency.

\paragraph{Topology-aware partitioning and dynamic load balancing for visual inputs.}
For visual inputs such as long videos, where sequence lengths vary significantly, we further introduce a topology-aware partitioning and dynamic load-balancing scheme. 
In a conventional implementation, partitioning is performed during the forward pass, which means each rank must first hold the full patch tensor before redistribution, leading to unnecessary memory and communication overhead.
To address this, we move CP and TP partitioning upstream into the data-loading stage and align partition boundaries with downsample groups, thereby eliminating the need for cross-rank patch aggregation. 
After load balancing across DP groups, precise dispatch is carried out through asynchronous all-to-all communication, so that each rank receives only the partition it actually needs. 
We further move large Python objects off the GPU communication path and onto the CPU path, which reduces GPU communication buffer overhead by about 7 GB in practice. 
For the variable-length sequences produced during rollout, we additionally perform joint bin-packing over both sequence length and ViT token count, leading to better-balanced micro-batches for both compute and memory pressure.

%% file: 3_multimodal_agentic.tex
\section{Multimodal Agent Capabilities and Ecosystem}
\subsection{Multimodal Toolchain Expansion}

\begin{table}[h]
\centering

\caption{Categorization of multimodal tools and processing functions based on application scenarios and tool sets. Tools prefixed with \texttt{zai\_} are proprietary developments, while the \textit{GLM-5V-Turbo} model also maintains compatibility with other user-defined custom tools.}

\begin{tabular}{lll}
\toprule
\textbf{Scenarios} & \textbf{Tool Sets} & \textbf{Tool Names} \\
\midrule
\multirow{16}{*}{\textbf{General}} & \multirow{3}{*}{Recognition Tools} & zai\_recognize\_plant \\
 &  & zai\_recognize\_location \\
 &  & zai\_recognize\_person \\
\cmidrule{2-3}
 & \multirow{5}{*}{Multimodal Search} & zai\_search\_web\_text \\
 &  & zai\_search\_web\_by\_image \\
 &  & zai\_search\_similar\_images \\
 &  & zai\_search\_web\_images \\
 &  & zai\_search\_scholar \\
\cmidrule{2-3}
 & \multirow{2}{*}{Browser Tools} & zai\_load\_image\_from\_url \\
 &  & zai\_read\_webpage \\
\cmidrule{2-3}
 & \multirow{6}{*}{Image Processing} & zai\_crop\_image \\
 &  & zai\_draw\_image\_bounding\_boxes \\
 &  & zai\_draw\_image\_point\_markers \\
 &  & zai\_draw\_image\_geometry \\
 &  & zai\_draw\_image\_3d\_bounding\_boxes \\
 &  & zai\_draw\_video\_objects\_tracking \\
\midrule
\multirow{6}{*}{\textbf{Creation}} & \multirow{4}{*}{Web Creation} & submit\_plan \\
 &  & apply\_edits \\
 &  & zai\_generate\_web\_html \\
 &  & zai\_generate\_web\_outline \\
\cmidrule{2-3}
 & \multirow{2}{*}{Slide Creation} & zai\_generate\_slide\_html \\
 &  & zai\_generate\_outline\_ppt \\
\midrule
\multirow{6}{*}{\textbf{Deep Research}} & \multirow{6}{*}{Multimodal DR Tools} & zai\_dr\_python \\
 &  & zai\_dr\_open\_url\_mm \\
 &  & zai\_dr\_visit\_img \\
 &  & zai\_dr\_search \\
 &  & zai\_dr\_images\_search \\
 &  & zai\_dr\_images\_lens \\
\bottomrule
\end{tabular}
\end{table}

GLM-5V-Turbo further expands its multimodal toolchain\footnote{The proprietary tools can be accessed and experienced through GLM-5V-Turbo model on \url{https://chat.z.ai/}.}, enabling the model to support a fuller perception–planning–execution loop in more realistic environments. In addition to expanding its repertoire of visual tools, the model demonstrates a sophisticated ability to maintain long-horizon engagement, frequently switching between multimodal search, annotation, screenshotting, and multimodal webpage reading tools to achieve thorough task resolution. Consequently, coding and task execution are no longer confined to textual interfaces but are instead iteratively grounded in a comprehensive, vision-based understanding of the environment.

These architectural advancements are validated by significant performance gains across specialized benchmarks. Compared to our recent model GLM-4.6V~\cite{vteam2025glm45vglm41vthinkingversatilemultimodal}, GLM-5V-Turbo demonstrates a substantial leap in complex multimodal tasks; notably, it achieves a score of 30.0 on MMSearch-Plus~\cite{MMSearch-Plus}, nearly an eightfold improvement over the previous generation. Strong growth is also evident in BrowseComp-VL~\cite{BrowsecompVL} (51.9) and ImageMining (30.7), which specifically test the model's ability to navigate web interfaces and extract deep visual insights. By matching or exceeding the performance of industry benchmarks like Kimi K-2.5~\cite{kimiteam2026kimik25visualagentic} and Claude Opus 4.6~\cite{anthropic2026claudeopus46} in these categories, GLM-5V-Turbo proves its capability to handle the high-dimensional reasoning required for modern agentic workflows.

This expansion is particularly important for multimodal agents. Many real-world tasks are not simply a matter of reading text and calling functions; they require the model to first interpret the visual environment, decide what to do next, and then continue adapting its behavior based on the outcome of its actions. For example, when reproducing a real website, the model can first use a multimodal GUI agent to explore the site through screenshots, interaction with page elements, and navigation across pages, building a richer understanding of layout, functionality, and interaction flow. It can then rely on its native UI-to-code capability to reproduce the site more faithfully. Likewise, when media assets such as images need to be incorporated, they can be processed directly through native tools such as cropping before being embedded into the final output.


\subsection{Integration with External Agent Frameworks: Claude Code and AutoClaw}

A critical component of GLM-5V-Turbo’s deployment strategy is its seamless integration with industry-standard external agent frameworks. By moving beyond isolated tool calls, the model serves as the cognitive core for systems like Claude Code and AutoClaw~\cite{zhipuai2026autoclaw}, bridging the gap between high-level reasoning and low-level system execution.
The integration with Claude Code transforms GLM-5V-Turbo from a passive code generator into an active system-level collaborator. Within this framework, the model leverages its multimodal capabilities to navigate complex terminal environments and local file systems. While Claude Code handles the logic and environment, AutoClaw provides the "hands" for browser-based and GUI-centric automation. GLM-5V-Turbo acts as the vision-language controller for AutoClaw, enabling sophisticated agentic workflows.

The convergence of GLM-5V-Turbo with these frameworks facilitates a complete perception–planning–execution loop. By offloading specific execution logic to Claude Code and AutoClaw, the model can focus on high-dimensional reasoning. This transition marks a fundamental shift in the model's role: it is no longer just a text-based assistant, but a multimodal actor grounded in real-world environments, capable of autonomous task resolution across diverse digital interfaces.

\subsection{ImageMining: A Self-Collected Vision-Centric Deep Search Benchmark}

The core potential of a multimodal agent lies in anchoring reasoning within visual contexts—a paradigm we term ``think with image, deep search with image.'' To evaluate this, we introduce \textbf{ImageMining}\footnote{\url{https://github.com/zai-org/ImageMining}}, a benchmark designed to test the integration of high-density visual understanding and autonomous multimodal search.

Unlike traditional VQA~\citep{BrowsecompVL, jiang2024mmsearch, MMSearch-Plus}, ImageMining requires models to actively mine visual inputs through agentic behaviors. Success relies on multi-step tool calls, such as localized cropping or magnification of minute details to refine search queries. This ``Deep-Wide-Search'' spectrum evaluates models on their search breadth across sources and their depth in visual reasoning, where task performance correlates strongly with the precision of on-image tool usage.

ImageMining comprises 217 curated test cases derived from manually collected trace samples, spanning seven domains (Social, Entertainment, Products, Places, Rich Text, Nature, and Science) and five reasoning categories:
\begin{itemize}[leftmargin=1.5em,itemsep=0pt,parsep=0.2em,topsep=0.1em,partopsep=0.0em]
\item \textbf{Universal Recognition:} Fine-grained identification of flora, fauna, and artifacts.
\item \textbf{Spatio-Temporal Reasoning:} Geographic deduction grounded in visual cues.
\item \textbf{Event Reasoning:} Comprehension of news events and product launches.
\item \textbf{Text-based Reasoning:} Reasoning over embedded rich text (e.g., academic papers, reports).
\item \textbf{Visual Search:} Cross-referencing visual inputs to retrieve specific artworks or imagery.
\end{itemize}

To equip GLM-5V-Turbo with these capabilities, we developed a multi-stage automated data pipeline covering knowledge discovery, QA reconstruction, and quality filtering. A pivotal constraint in this process is the \textbf{``Visual Jump''} (\texttt{WEB\_VISUAL}): during discovery, intermediate reasoning hops must involve visual transitions, forcing the model to parse images rather than relying on textual shortcuts or parametric knowledge. Furthermore, we constructed specialized \textbf{OCR Search data} for charts, maps, and posters. This compels the model to perform entity isolation and localized cropping before initiating search chains, transforming images from static inputs into interactive environments for deep exploration.

\subsection{Multimodal Deep Research and Content Creation}

\begin{figure}[htbp]
    \centering
    \begin{subfigure}{0.492\textwidth}
        \includegraphics[width=\textwidth]{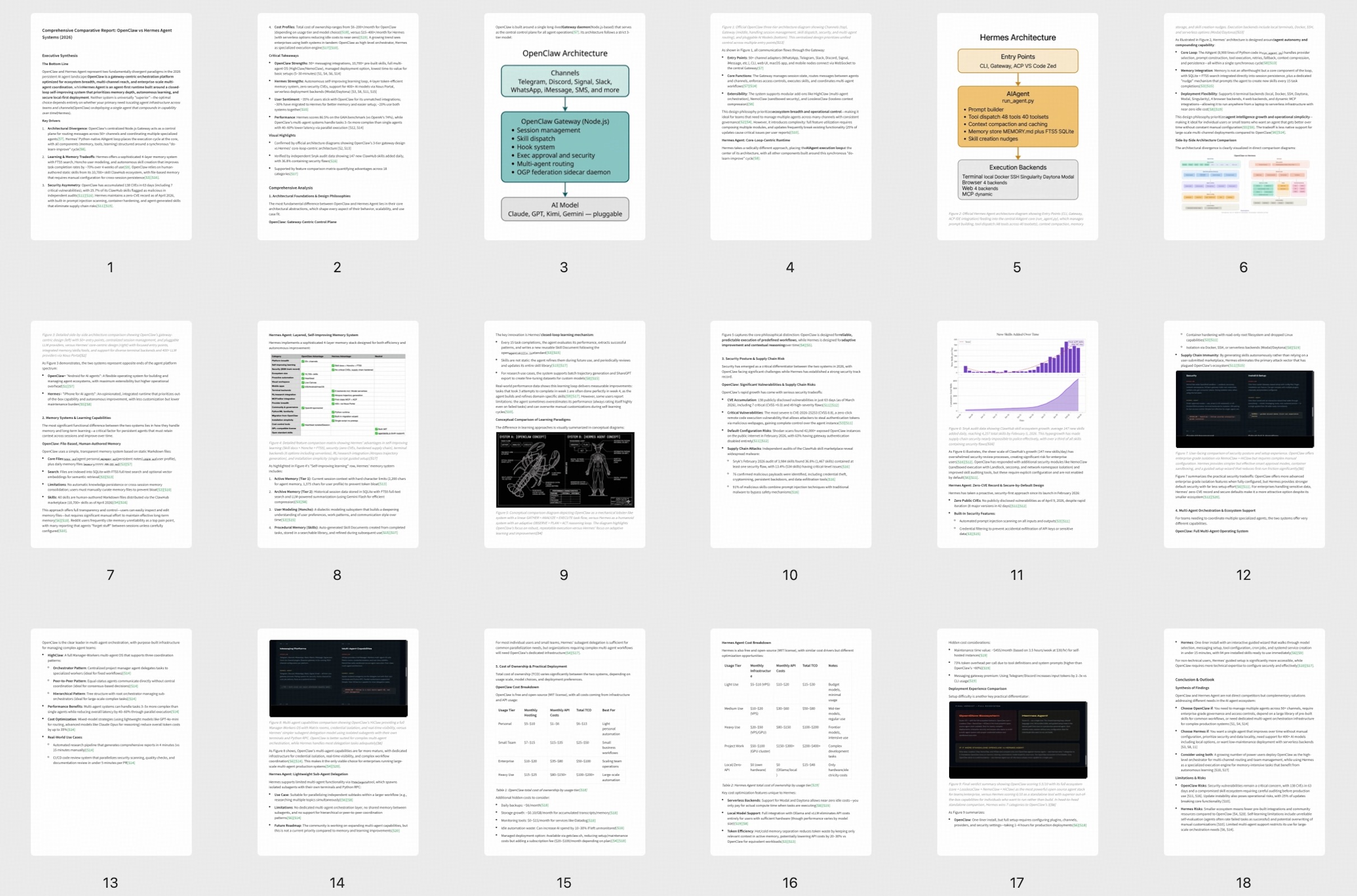}
        \caption{}
        \label{fig:sub1}
    \end{subfigure}
    \hfill
    \begin{subfigure}{0.473\textwidth}
        \includegraphics[width=\textwidth]{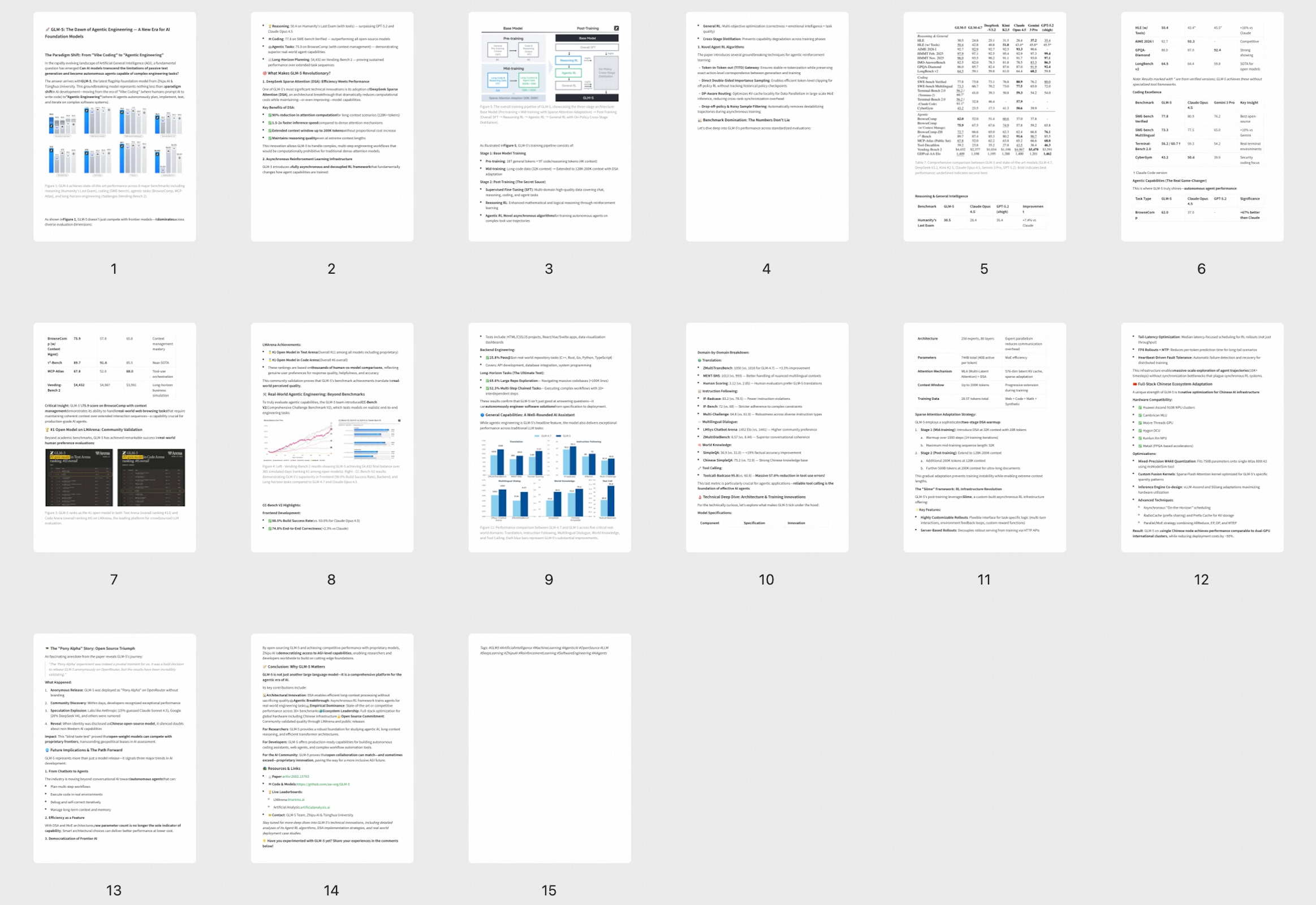}
        \caption{}
        \label{fig:sub2}
    \end{subfigure}
    \caption{Examples of multimodal deep research and content creation. (a) A multimodal deep research report, where the visuals are harvested from the Internet via web search, and selected and complied by GLM-5V-Turbo (Query: \textit{Compare OpenClaw and Hermes agent systems and give a comprehensive report. Note that the output should be a text-image interleaved markdown.}). (b) A technical blog excerpted from an academic paper \cite{zeng2026glm}, where the visual elements are cropped from the original paper and inserted into the output to compose a complete blog, fully automated by GLM-5V-Turbo.}
    \label{fig:mmdr}
\end{figure}

Leveraging its agentic capabilities, GLM-5V-Turbo facilitates a complete multimodal deep research workflow, encompassing iterative information gathering, evidence consolidation, and long-form synthesis from heterogeneous sources. Unlike traditional text-centric agents~\cite{google_gemini_deep_research_2025, openai_deep_research_2025}, this workflow begins with open-ended objectives and proceeds through autonomous cycles of planning, multimodal reading, and state updating. By natively parsing visually rich webpages, charts, and structured documents, the model accesses high-value evidence—such as slides and figures—that is typically discarded in text-only pipelines.

A defining characteristic of this system is its integrated multimodal reasoning. Rather than treating images as peripheral data, GLM-5V-Turbo extracts textual and visual evidence (e.g., table regions, screenshots) in tandem. This is crucial for realistic research environments where key insights are often distributed across document layouts and visual artifacts rather than isolated within text paragraphs.

Beyond information acquisition, GLM-5V-Turbo supports diverse, presentation-oriented downstream formats:
\begin{itemize}[leftmargin=1.5em,itemsep=0pt,parsep=0.2em,topsep=0.1em]
\item \textbf{Interleaved Reports:} Generating text-image interleaved outputs (see Fig.~\ref{fig:mmdr} (a)) where visual evidence is embedded alongside grounded explanations—ideal for comparative analysis and literature reviews.
\item \textbf{Deep Research to PPT:} Synthesizing gathered materials into structured slide decks, including page allocation and multimodal content organization, to mirror professional presentation workflows.
\item \textbf{Document-Style Write-ups:} Creating blog-like interpretations or structured notes (see Fig.~\ref{fig:mmdr} (b)) that maintain the visual-textual integrity of the research findings.
\end{itemize}

These capabilities further extend to document-grounded generation. Users can provide complex source materials for the model to reorganize into structured slides or interleaved interpretations. By preserving the synergy between textual conclusions and supporting visual evidence, GLM-5V-Turbo marks a system-level transition from simple multimodal information retrieval to comprehensive multimodal transformation and presentation.


\subsection{Official Skills}

\begin{table}[htp]
  \centering
  \caption{Overview of official skills supported by GLM-5V-Turbo.}
  \label{tab:official_skills}
  \begin{tabularx}{\textwidth}{c c X}
    \toprule
    \textbf{Skill} & \textbf{Type} & \textbf{URL} \\
    \midrule
    \scriptsize{\texttt{PDF-to-Web}} & \scriptsize{Native} & \scriptsize{\url{https://clawhub.ai/zai-org/glmv-pdf-to-web}} \\
    \scriptsize{\texttt{PDF-to-PPT}} & \scriptsize{Native} & \scriptsize{\url{https://clawhub.ai/zai-org/glmv-pdf-to-ppt}} \\
    \scriptsize{\texttt{Web Replication}} & \scriptsize{Native} & \scriptsize{\url{https://clawhub.ai/zai-org/glmv-web-replication}} \\ 
    \scriptsize{\texttt{PRD-to-App}} & \scriptsize{Native} & \scriptsize{\url{https://clawhub.ai/zai-org/glmv-prd-to-app}} \\
     \scriptsize{\texttt{Stock Analyst}} & \scriptsize{Native} & \scriptsize{\url{https://clawhub.ai/zai-org/glmv-stock-analyst}} \\
    \scriptsize{\texttt{Image Captioning}} & \scriptsize{External Tool} & \scriptsize{\url{https://clawhub.ai/JaredforReal/glmv-caption}} \\
    \scriptsize{\texttt{Visual Grounding}} & \scriptsize{External Tool} & \scriptsize{\url{https://clawhub.ai/jaredforreal/glmv-grounding}} \\
    \scriptsize{\texttt{Doc-based Writing}} & \scriptsize{External Tool} & \scriptsize{\url{https://clawhub.ai/jaredforreal/glmv-doc-based-writing}} \\
    \scriptsize{\texttt{Resume Screening}} & \scriptsize{External Tool} & \scriptsize{\url{https://clawhub.ai/JaredforReal/glmv-resume-screen}} \\
    \scriptsize{\texttt{Prompt Generation}} & \scriptsize{External Tool} & \scriptsize{\url{https://clawhub.ai/JaredforReal/glmv-prompt-gen}} \\
    \scriptsize{\texttt{General OCR}} & \scriptsize{Specialized} & \scriptsize{\url{https://clawhub.ai/JaredforReal/glmocr}} \\
    \scriptsize{\texttt{Table Recognition}} & \scriptsize{Specialized} & \scriptsize{\url{https://clawhub.ai/JaredforReal/glmocr-table}} \\
    \scriptsize{\texttt{Handwriting Recognition}} & \scriptsize{Specialized} & \scriptsize{\url{https://clawhub.ai/JaredforReal/glmocr-handwriting}} \\
    \scriptsize{\texttt{Formula Recognition}} & \scriptsize{Specialized} & \scriptsize{\url{https://clawhub.ai/JaredforReal/glmocr-formula}} \\
    \scriptsize{\texttt{Image Generation}} & \scriptsize{Specialized} & \scriptsize{\url{https://clawhub.ai/JaredforReal/glm-image-gen}} \\
    \bottomrule
  \end{tabularx}
\end{table}

As a foundation model adept at agentic and coding tasks, GLM-5V-Turbo can be readily integrated into general and coding agent frameworks (such as OpenClaw~\cite{steinberger2026openclaw}, AutoClaw~\cite{zhipuai2026autoclaw} and Claude Code~\cite{anthropic2025claudecode}), which are becoming increasingly popular in the community. To make it easier for users to utilize GLM-5V-Turbo within these agent systems, and to better leverage its strengths, we provide a set of official skills, which fall into two categories: one is built upon the native capabilities of the GLM-5V-Turbo model, and the other wraps GLM-5V-Turbo as an external tool (in the form of a MaaS API) for OpenClaw, AutoClaw and Claude Code to invoke. Additionally, we have developed 5 skills based on the previously released specialized models, GLM-OCR~\cite{duan2026glmocrtechnicalreport} and GLM-Image~\cite{zhipuai2026glmimage}, to support a wider range of scenarios and tasks. To help users better understand, install, and use the official skills, we also provide a unified master skill (\url{https://clawhub.ai/jaredforreal/glm-master-skill}).

The official skills are listed in Tab.~\ref{tab:official_skills} and more details can be found in the Github repository: \url{https://github.com/zai-org/GLM-skills}.

%% file: 4_design_lenses.tex
\section{Design Lenses from Development}
Beyond the developments described above, the process of building GLM-5V-Turbo also led us to several practical lenses for agentic model development. We present them not as universal rules, but as design perspectives that repeatedly proved useful in our development process.

\begin{lensbox}
\textbf{Lens 1:} Perception remains foundational to higher-level multimodal capability.
\end{lensbox}

Recent work has placed increasing emphasis on higher-level abilities such as planning, reasoning, and reflection. Our observation, however, is that further gains in multimodal capability still depend critically on perception. Even among the strongest current VLMs, errors in fine-grained perception and spatial understanding remain common, and these often propagate into downstream reasoning, decision-making, and execution. Many failures that appear high-level, in other words, begin with the model not seeing the environment accurately enough.

In our development, multimodal coding and grounding proved to be useful proxy tasks for perceptual learning. Tasks such as frontend or SVG coding require the model to capture layout, structure, relative position, and local detail, rather than relying only on coarse semantics. 
We found that adding paired data between subject-specific images and their SVG representations during pretraining contributed positively to downstream STEM problem solving, while strengthening grounding-related training during RL also improved GUI-agent performance. These observations suggest that some seemingly downstream structured tasks can in fact provide a useful route to better perception.

We also find that explicitly training the model to critique its own perception can help reduce hallucination during generation. In GUI-agent instruction tuning, we include a subset of critic data that targets errors in the reasoning process, such as misreading interface details, misidentifying target elements, and making incorrect decisions about the next action. This improves the model’s observation quality on GUI details and reduces several recurring perception failure modes. 
More broadly, our view is that perception is not a low-level module that can simply be solved early and then left behind; it continues to shape the upper bound of higher-level multimodal capability.

\begin{lensbox}
\textbf{Lens 2:} Agent capability can be more efficiently built through hierarchical optimization.
\end{lensbox}

Agent training is inherently resource-intensive: environment setup and task construction are costly, high-quality data is scarce, and reliable verification is often difficult. 
At the same time, agent tasks themselves are hard to optimize efficiently, since they typically involve complex compositions, long interaction trajectories, non-unique solution paths, and strong dependence on the evolving environment state. 
Under these conditions, a central question is how to maximize the return on data construction under limited resources.

This led us to adopt a hierarchical optimization strategy. In our experience, agent capability is developed more effectively when optimization is distributed across multiple levels of the capability hierarchy, rather than concentrated primarily on high-level long-horizon tasks. 
In GUI-agent development, for example, this motivated us to build a multi-level task hierarchy spanning element perception, GUI grounding, single-step action prediction, and trajectory-level action prediction, and to use it in both SFT and RL. 
The appeal of this design is twofold: lower-level tasks are usually easier to construct, annotate, and verify than long-horizon ones under the same resource constraints; and when lower-level capabilities are still underdeveloped, pushing only on high-level tasks often fails to yield reliable gains and can instead make training less stable. 
Overall, hierarchical optimization serves not only as a way to improve efficiency, but also as a practical path toward more stable agent training.

\begin{lensbox}
\textbf{Lens 3:}  The key to constructing, evaluating, and optimizing end-to-end long-horizon tasks lies in clear task specification, reliable outcome verification, and controlled evaluation procedures.
\end{lensbox}

For multimodal agents, the real challenge is often not extending tasks to longer horizons, but making end-to-end tasks stable enough to serve as meaningful targets for evaluation and optimization. Many realistic agent settings are inherently open-ended, with underspecified goals, ambiguous execution boundaries, and outcomes that depend heavily on intermediate decisions. As a result, they are often difficult to compare consistently and even harder to turn into reusable optimization signals.

This led us to a broader view: the value of an end-to-end task depends not only on how realistic it is, but also on whether it can be specified clearly enough, verified reliably enough, and evaluated under sufficient procedural control to produce stable and reusable feedback. 
This perspective shaped how we think about data construction, evaluation, and downstream optimization. In multimodal agent settings, task definition often depends on multiple sources of constraint rather than a single prompt alone, while evaluation needs structure not only at the level of final outcomes but also at the level of the verification process itself. Under this view, task definition, verification design, and feedback structure should be considered together rather than in isolation.

Vision2Web~\cite{he2026vision2web}, our benchmark for end-to-end visual website development, is one concrete instantiation of this view. Each task is grounded not just in a textual instruction, but in a richer specification that may include PRDs, mockups, reference pages, and resource assets, making the task definition better specified. 
On the evaluation side, rather than treating website development as a loosely specified open-ended problem, we use workflow-based verification so that execution is assessed through a controlled sequence of dependent steps rather than a single final state. 
This makes it easier to compare systems, attribute failures, and model different forms of signal separately --- for example, functional correctness during interactive execution and visual consistency in a more isolated comparison setting. In this sense, Vision2Web is not only a benchmark, but also a concrete attempt to align task construction, verification, and feedback design in a way that better supports reliable evaluation and optimization.

%% file: 5_evaluation.tex
\section{Evaluation}
We evaluate GLM-5V-Turbo across four categories: 

\begin{itemize} [leftmargin=1.5em,itemsep=0pt,parsep=0.2em,topsep=0.1em,partopsep=0.0em]
    \item \textbf{Multimodal Coding}:
        Desing2Code~\cite{si2025design2code},
        Flame-VLM-Code~\cite{ge2025advancing},
        Vision2Web~\cite{he2026vision2web};
    \item \textbf{Multimodal ToolUse}:
        ImageMining,
        BrowseComp-VL~\cite{BrowsecompVL},
        MMSearch~\cite{MMSearch},
        MMSearch-Plus~\cite{MMSearch-Plus},
        SimpleVQA~\cite{SimpleVQA},
        Facts~\cite{FACTS},
        V*~\cite{wu2023vguidedvisualsearch};
    \item \textbf{GUI Agent}:
        OSWorld~\cite{xie2025osworld}, AndroidWorld~\cite{rawles2024androidworld},
        WebVoyager~\cite{he2024webvoyager};
    \item \textbf{Text-only Coding and Claw}:
        CC-Bench-V2~\cite{zeng2026glm},
        PinchBench~\cite{PinchBench},
        ClawEval~\cite{ClawEval},
        ZClawBench~\cite{ZClawBench}.
\end{itemize}

Across these dimensions, GLM-5V-Turbo exhibits a consistent pattern: it achieves strong performance on multimodal benchmarks for coding and agent-oriented tasks, while maintaining solid capability on text-only tasks. This balance aligns with our core objective for GLM-5V-Turbo: \textbf{building foundational multimodal agentic capability without sacrificing the coding and reasoning ability required in text-first workflows.}

\begin{figure}[h]
    \centering
    \includegraphics[width=0.75\linewidth]{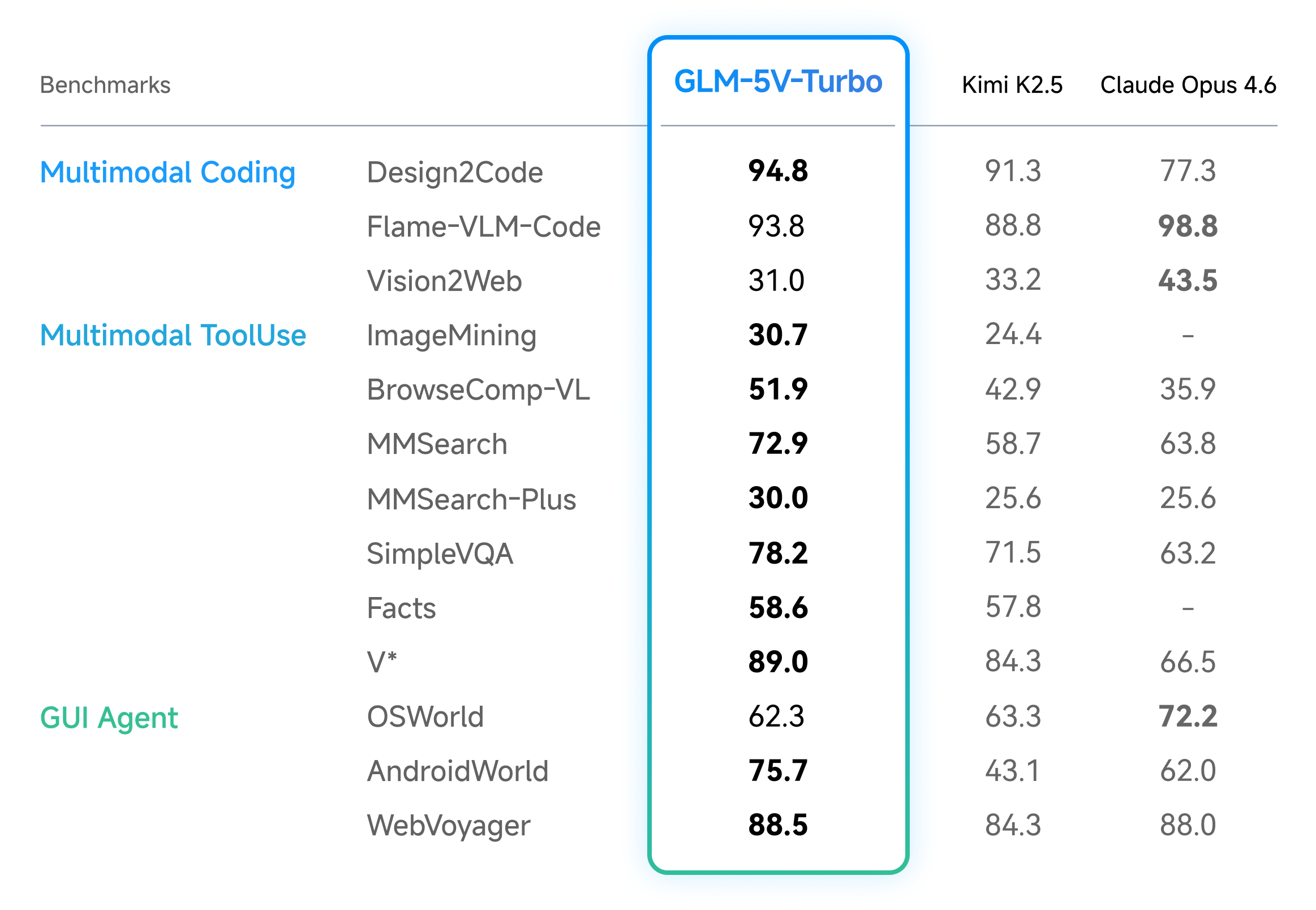}
    \caption{Evaluation of GLM-5V-Turbo on multimodal coding, tool-use, and GUI agent benchmarks. }
    \label{fig:benchmark1}
\end{figure}

On multimodal coding and tool-use benchmarks, GLM-5V-Turbo performs strongly on UI-to-code generation, visual website development, multimodal search, and visually grounded QA. It is also highly competitive on GUI-agent benchmarks such as AndroidWorld and WebVoyager, indicating that its visual understanding transfers effectively into grounded interaction and action. 
At the same time, on CC-Bench-V2 including CC-Backend, CC-Frontend, and CC-Repo-Exploration  which evaluate model performance on Claude Code framework, the model remains solid in pure-text coding, suggesting that the addition of visual capability does not materially erode its underlying coding performance, which is a critical feature for the multimodal agentic foundations. 

\begin{figure}[h]
    \centering
    \includegraphics[width=0.75\linewidth]{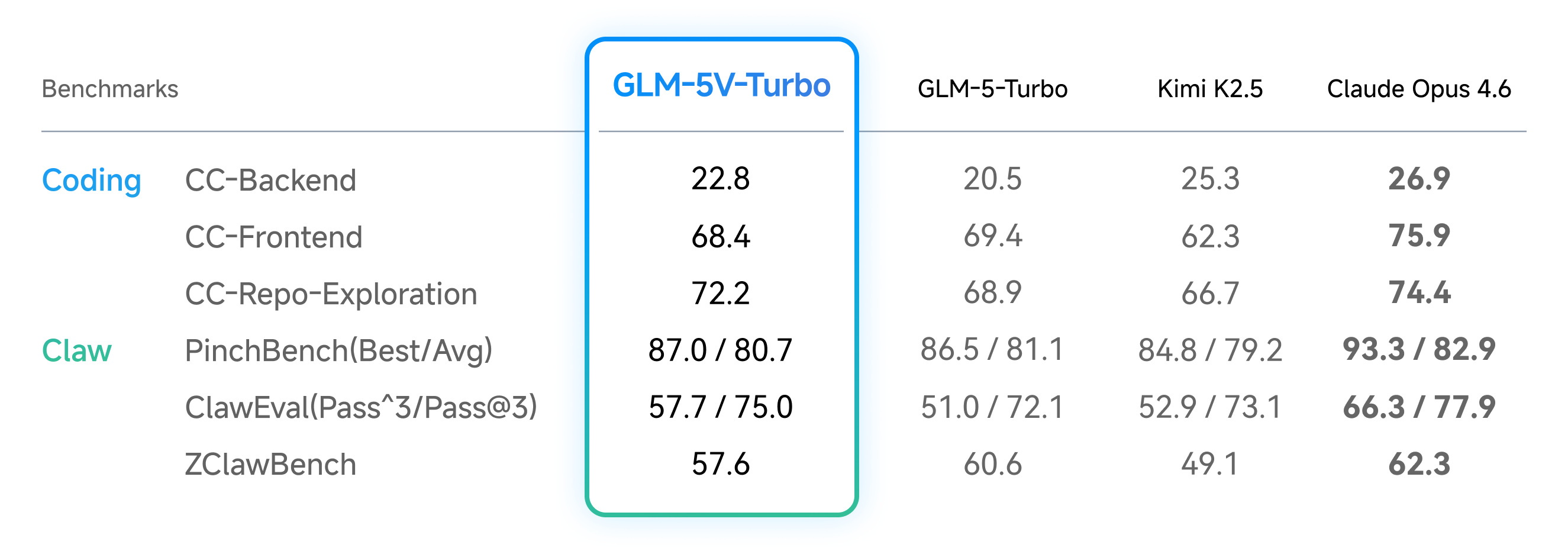}
    \caption{Evaluation of GLM-5V-Turbo on text coding and claw agent benchmarks. }
    \label{fig:benchmark2}
\end{figure}

We also find that GLM-5V-Turbo transfers effectively to vision-enabled general agent frameworks. In particular, when integrated into Claw agent frameworks, the model can natively perceive on-screen content and act on it more effectively, leading to strong results on execution-oriented evaluations such as PinchBench, ClawEval, and ZClawBench. While Claw is only one representative framework, these results provide further evidence that the model’s multimodal capability is not limited to isolated benchmark gains, but carries over to realistic end-to-end agent execution.

%% file: 6_remaining_challenges.tex
\section{Remaining Challenges}
Despite the progress described above, several challenges remain central to future agentic model development. In our view, the hardest open problems increasingly lie not in isolated capability improvement, but in agentic strategy emergence, long-horizon multimodal context management, and the growing entanglement between model capability and harness design.

\paragraph{How to enable the emergence of better agentic strategies.}
Agent training still depends heavily on hand-crafted or strongly filtered cold-start trajectories. This is effective for initialization, but it also narrows the space of reasoning and action patterns the model is likely to explore, so later improvement often remains local: the model becomes better at executing familiar paths, without discovering genuinely better ones. 
In our experiments, we found that increasing trajectory diversity at the cold-start stage can partially loosen this constraint, making it easier for RL to uncover nearby but improved variants. 
This suggests that trajectory diversity is not merely a matter of broader data coverage, but may be one of the conditions for strategy emergence itself. Still, this is only a first step. The more fundamental goal is to enable models to discover better reasoning and agentic strategies on their own, rather than remaining confined to variations of human-provided starting patterns. Beyond that lies an even harder challenge: enabling models to discover richer organizational forms, such as sub-agent decomposition, multi-agent collaboration, and more flexible hierarchical decision structures.

\paragraph{Multimodal context management remains a core bottleneck for long-horizon agents.}
Compared with text, images and especially videos consume context budget much more aggressively, making them expensive to retain over long trajectories. In practice, many systems respond by dropping earlier visual observations as context grows. While being an understandable engineering compromise, it also discards information that may remain important for later reasoning, planning, or verification. 
The challenge becomes sharper as trajectories lengthen. 
In text-only settings, systems such as Claude Code often respond to growing context pressure by compacting or summarizing earlier interaction history once the context window starts to fill up; in multimodal settings, however, faithful compression is much harder, because what must be preserved is not only semantic content, but also visual detail that may later become important again, such as layout, spatial relations, or temporal change in video. 
Most current memory mechanisms remain fundamentally text-centric: they are better at compressing what was said than what was seen, or how visual states evolved over time. For long-horizon multimodal agents, simply adapting text memory mechanisms will therefore be insufficient. What is needed instead is a more multimodal-native approach to context and memory.

\paragraph{Model and harness increasingly co-shape the system’s capability boundary.}
For agentic systems, the effective capability boundary is no longer determined by the model alone, but jointly shaped by the model and the harness around it. 
This greatly expands the design space: task decomposition, tool use, memory mechanisms, and verification loops can all affect what the system is able to do in practice. 
At the same time, it makes the development path substantially complex: the same model may behave very differently under different decomposition strategies, tool-use policies, memory designs, or verification workflows; conversely, what appears to be a model limitation may sometimes reflect a poor harness choice instead. 
More importantly, this dependence runs both ways: the usefulness of a harness often depends on the model’s capability regime, and designs that are ineffective at one stage may become critical once the model crosses a threshold in reasoning, planning, or feedback utilization. 
This means the harness is not a stable external layer that can be optimized independently of the model. Its role, value, and optimal form shift as the model evolves. 
More broadly, this means that agentic model development can no longer be framed as model improvement alone: the effective capability boundary is increasingly co-shaped by the model and the harness, and so too are the objectives by which progress is optimized and evaluated.

%% file: 7_contributors.tex
\section{Contribution}
\label{sec:contribution}
\newcommand{\cpara}[1]{~\\\textbf{#1}~\\}
The contributors' names are listed in reverse alphabetical order (Z to A) by first name.

\cpara{Core Contributors}
Ziyang Pan, Zhen Yang, Yuting Wang, Yue Wang, Yuanchang Yue, Yu Wang,  Yanling Wang, Yan Wang, Xijun Liu, Wenmeng Yu, Weihan Wang, Wei Li, Shuaiqi Duan, Sheng Yang, Ruiliang Lv, Mingdao Liu, Lihang Pan, Ke Ning, Junhui Ji, Jinjiang Wang, Jing Chen, Jiazheng Xu, Jiale Zhu,  Jiale Cheng, Ji Qi, Guobing Gan, Guo Wang, Cong Yao

\cpara{Contributors}
Zijun Dou, Zihao Zhou, Zihan Wang, Zhiqi Ge, Zhijie Li, Zhenyu Hou, Zhao Xue, Zehui Wang, Zehan Qi, Zehai He, Yutao Zhang, Yusen Liu, Yukuo Cen, Yuchen Li, Yuan Wang, Yu Yang, Yongbin Liu, Yijian Lu, Yifan Xu, Yanzi Wang, Yanxiao Zhao, Yanfeng Wang, Yadong Xue, Yabo Xu,  Xinyu Zhang, Xinyu Liu, Xiao Liu, Wenyi Zhao, Wenkai Li, Tianyu Tong,  Tianshu Zhang, Shudan Zhang,  Shengdong Yan, Qinkai Zheng,  Mingde Xu, Licheng Bao, lat Long long, Jiaxing Xu, Jiaxin Fan, Jiawen Qian, Jiali Chen,  Jiahui Lin, Jiadai Sun, Haozhi Zheng, Haoran Wang,  Haochen Li, Hanyu Lai, Han Xu, Fan Yang, Dan Zhang, Da Yin, Chuangxin Zhao, Chengcheng Wu, Boyan Shi, Bowen Lv, Bowei Jia, Bo Li, Bin Chen, Baoxu Wang

\cpara{Tech Leads}
Wenyi Hong, Xiaotao Gu

\cpara{Academic Advisors}
Peng Zhang, Debing Liu, Bin Xu, Juanzi Li, Minlie Huang, Yuxiao Dong, Jie Tang

%% file: 8_appendix.tex
\section{Demo Cases}

We demonstrate the capabilities and advantages of GLM-5V-Turbo through typical qualitative examples from various scenarios.

\subsection{In Combination with Agent Systems and Skills}

\begin{figure}[!h]
    \centering
    \includegraphics[width=0.65\textwidth]{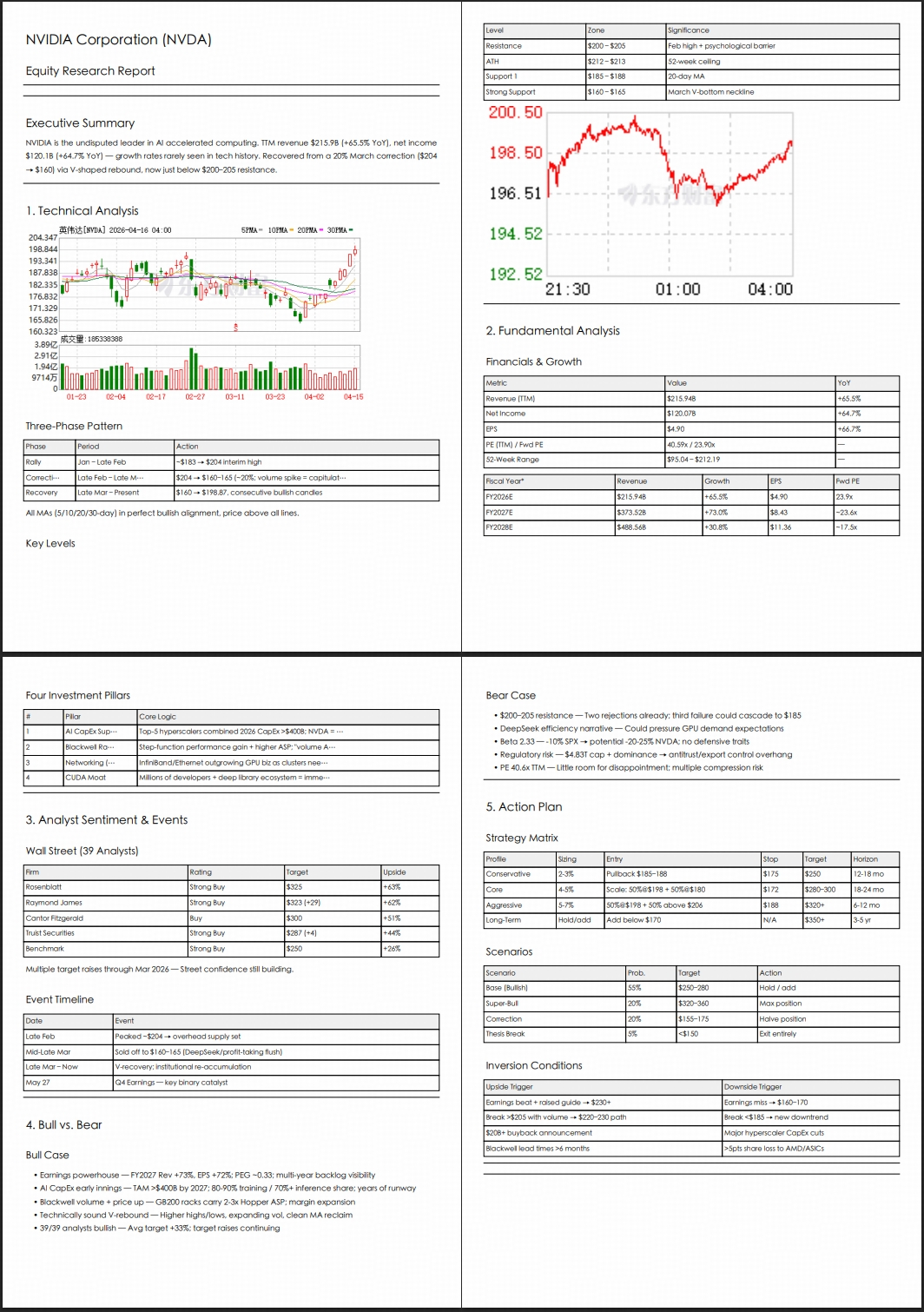} 
    \caption{A case showing the application of GLM-5V-Turbo to stock analysis, with OpenClaw and the official skill \texttt{glmv-stock-analyst}\protect\footnotemark. It gathers relevant information from multiple sources and produces a professional analysis report, including technical analysis, fundamental analysis, analyst sentiment and action plan. \textbf{Query:} \textit{Analyze NVIDIA's stock and give a English report.}}
    \label{fig:stock_analysis}
\end{figure}

\footnotetext{URL: \url{https://clawhub.ai/zai-org/glmv-stock-analyst}}

\clearpage



\begin{figure}[!h]
    \centering
    \includegraphics[width=0.95\textwidth]{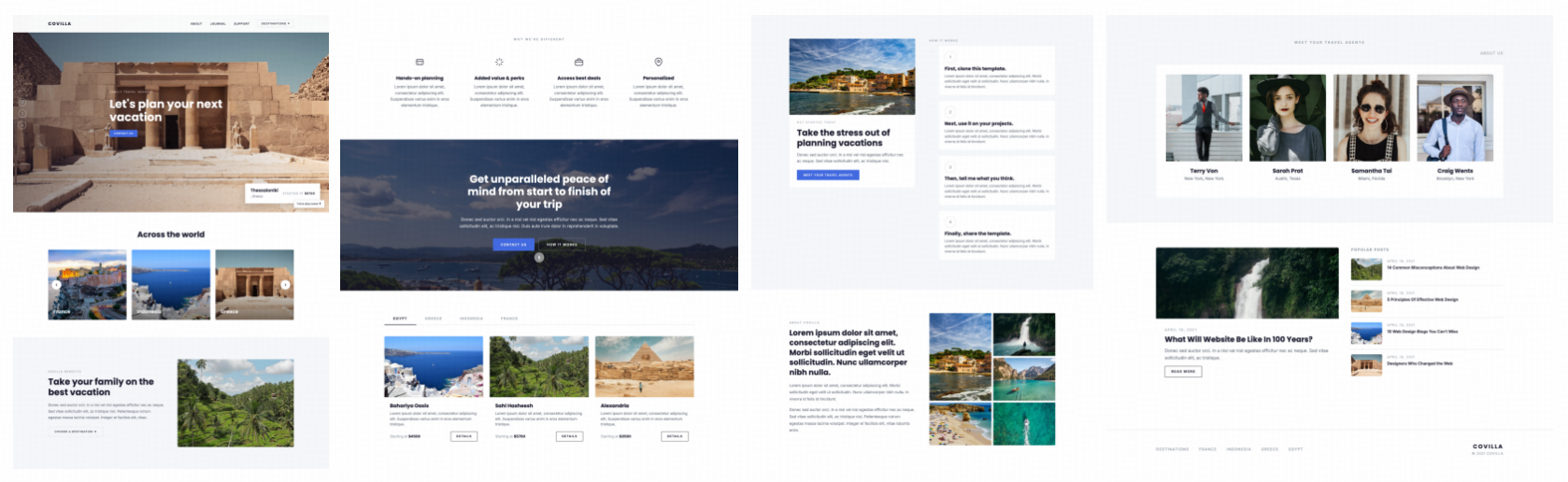}
    \caption{
    A case showing the application of GLM-5V-Turbo to URL-based GUI exploration, asset collection, and webpage recreation, with Claude Code and the official skill \texttt{glmv-web-replication}\protect\footnotemark.
    \textbf{Query:} \textit{Given a target website URL: \protect\url{https://webflow-path-three.webflow.io/}, please explore it via GUI, collect the necessary assets, and recreate the webpage in HTML code with high visual fidelity and functional completeness.}
    }
    \label{fig:mmgui_explore}
\end{figure}

\footnotetext{URL: \url{https://clawhub.ai/zai-org/glmv-web-replication}}


\begin{figure}[!h]
    \centering
    \includegraphics[width=0.95\textwidth]{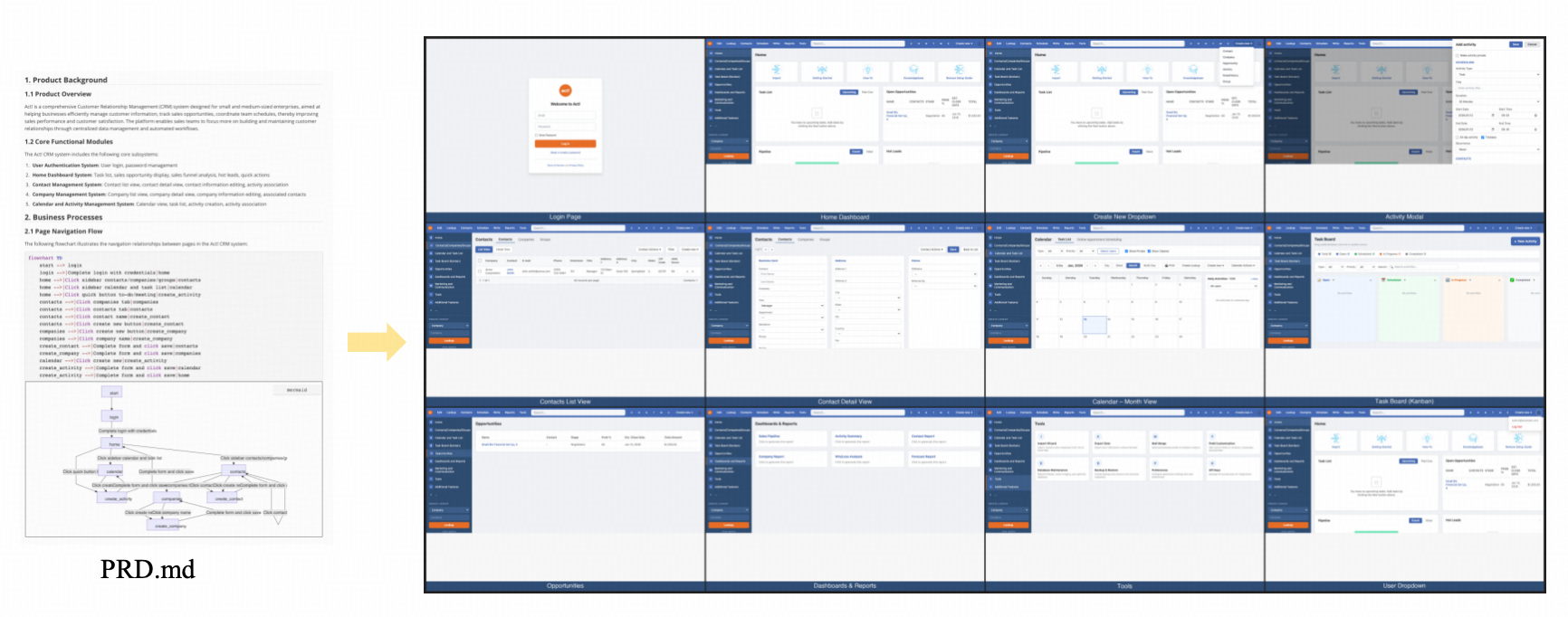} 
    \caption{A case showing the application of GLM-5V-Turbo to PRD-driven website generation, with Claude Code and the official skill \texttt{glmv-prd-to-app}\protect\footnotemark. Given a product requirements document and the project contents under the \texttt{act} folder, the model uses the PRD skill to design and implement a website in the working directory \texttt{./act\_workspace}. \textbf{Query:} \textit{Based on my PRD document, please use your PRD skills to build a website for the project in the \texttt{act} folder. The working directory is \texttt{./act\_workspace}.}}
    \label{fig:mm_prd_to_code}
\end{figure}

\footnotetext{URL: \url{https://clawhub.ai/zai-org/glmv-prd-to-app}}

\clearpage

\subsection{Multimodal Coding}


\begin{figure}[!h]
    \centering
    \includegraphics[width=0.95\textwidth]{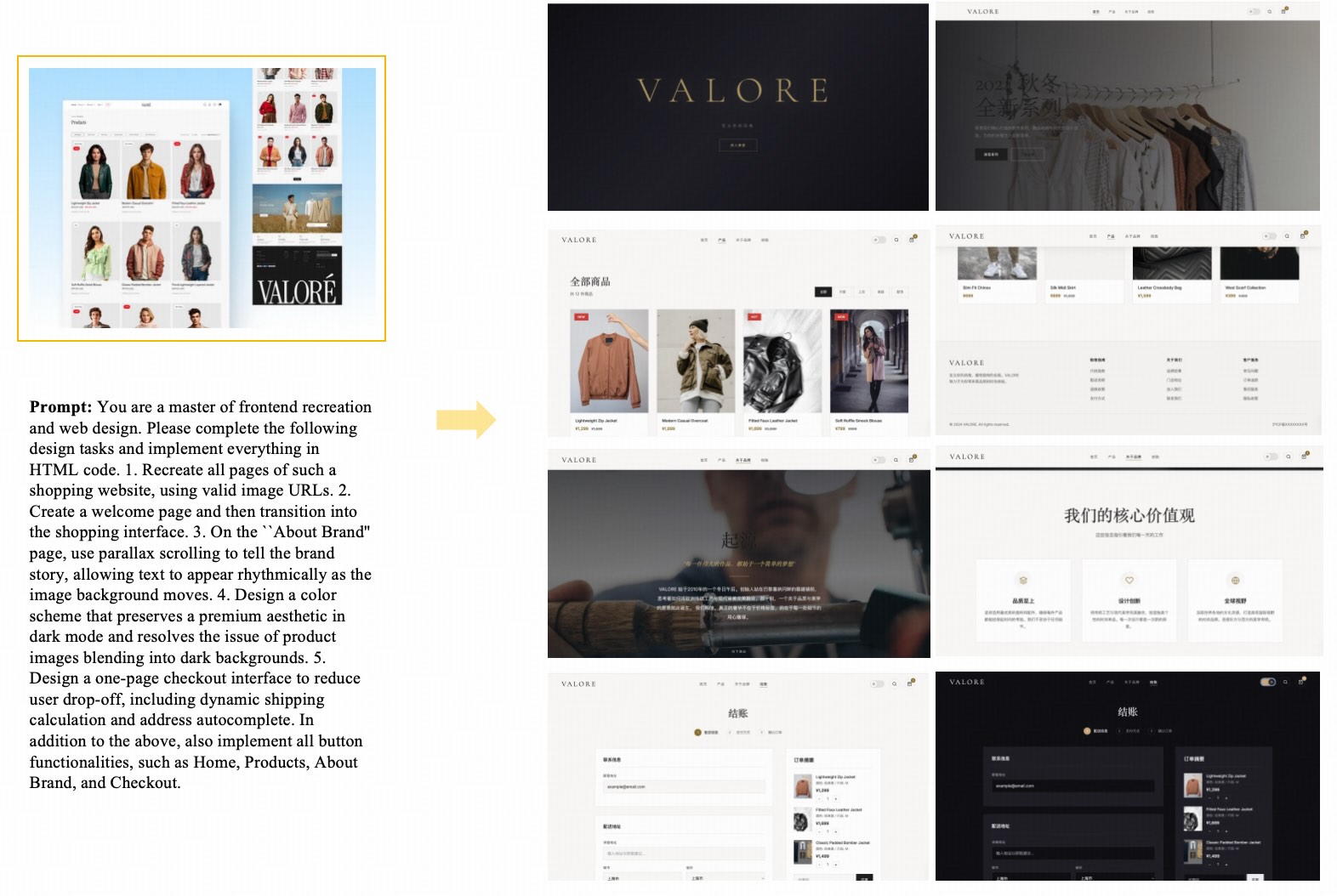} 
    \caption{A case showing the application of GLM-5V-Turbo to full-stack e-commerce website design and implementation, using our official website \textbf{z.ai}~\protect\footnotemark. Given a high-level product design request, the model generates a complete HTML-based shopping website with multiple functional pages, including a welcome page, shopping interface, brand-story page with parallax scrolling, dark-mode visual design, and a one-page checkout interface with dynamic shipping calculation and address suggestion. The model also completes interactive button behaviors across key pages such as Home, Products, About Brand, and Checkout. \textbf{Query:} \textit{You are a master of frontend recreation and web design. Please complete the following design tasks and implement everything in HTML code. 1. Recreate all pages of such a shopping website, using valid image URLs. 2. Create a welcome page and then transition into the shopping interface. 3. On the ``About Brand'' page, use parallax scrolling to tell the brand story, allowing text to appear rhythmically as the image background moves. 4. Design a color scheme that preserves a premium aesthetic in dark mode and resolves the issue of product images blending into dark backgrounds. 5. Design a one-page checkout interface to reduce user drop-off, including dynamic shipping calculation and address autocomplete. In addition to the above, also implement all button functionalities, such as Home, Products, About Brand, and Checkout.}}    
    \label{fig:mm_ui2code_web}
\end{figure}

\footnotetext{URL: \url{https://chat.z.ai/}}


\begin{figure}[!h]
    \centering
    \includegraphics[width=0.95\textwidth]{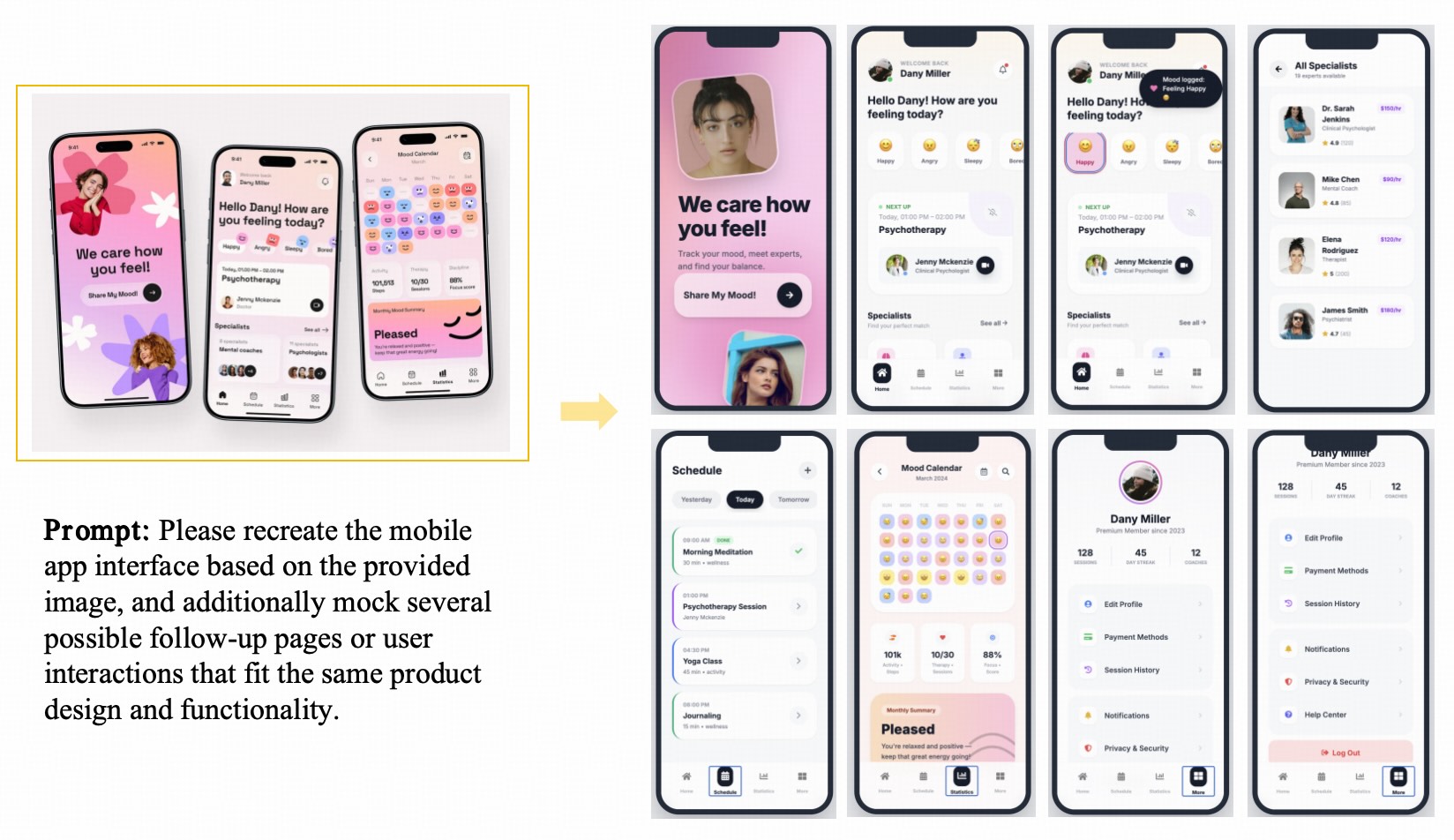} 
    \caption{A case showing the application of GLM-5V-Turbo to UI recreation and mock interface generation, using our official website \textbf{z.ai}. Given a reference image of a mobile mood-tracking application, the model reconstructs the interface in executable web code and further mocks additional plausible pages and interactions in a consistent visual style. \textbf{Query:} \textit{Please recreate the mobile app interface based on the provided image, and additionally mock several possible follow-up pages or user interactions that fit the same product design and functionality.}}
    \label{fig:mmui2code_mobile}
\end{figure}


\begin{figure}[!h]
    \centering
    \includegraphics[width=0.95\textwidth]{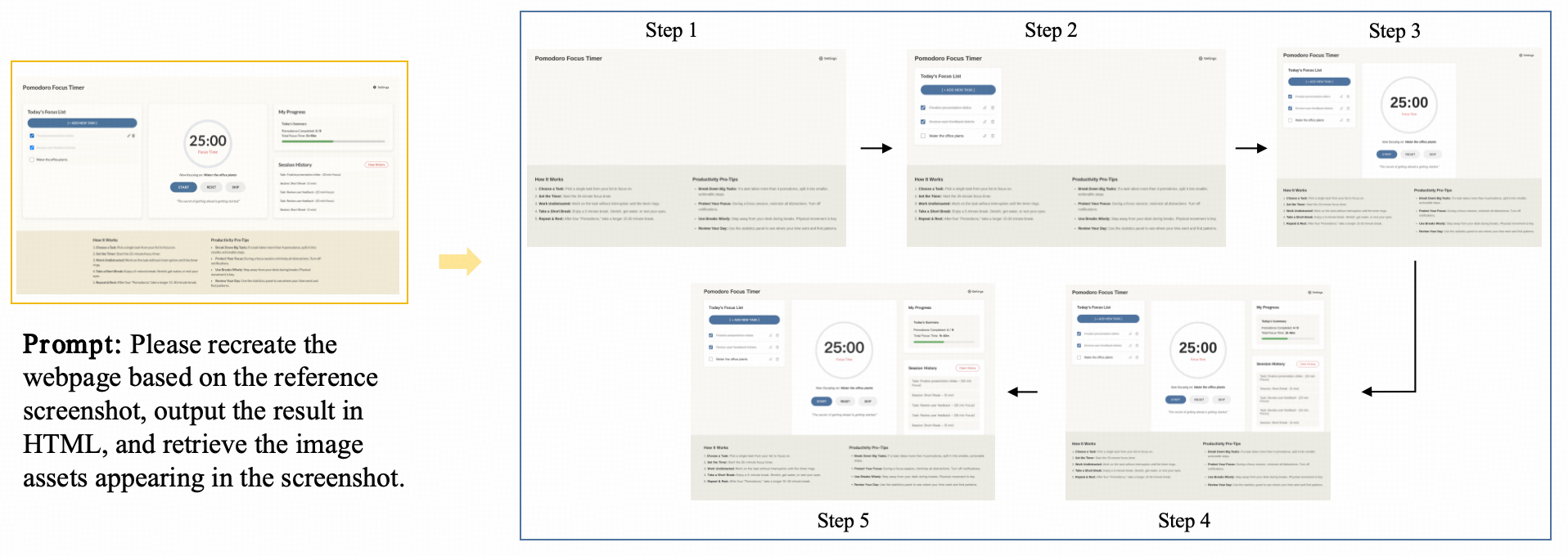} 
    \caption{A case showing the application of GLM-5V-Turbo to agentic UI recreation, using our official website \textbf{z.ai}. Given a reference screenshot of a webpage, the model reconstructs the page in HTML while automatically retrieving the image assets appearing in the screenshot. This example highlights the agentic framework's ability to jointly perform visual understanding, asset collection, and faithful UI recreation. \textbf{Query:} \textit{Please recreate the webpage based on the reference screenshot, output the result in HTML, and retrieve the image assets appearing in the screenshot.}}

    \label{fig:mm_agentic_ui2code}
\end{figure}


\begin{figure}[!h]
    \centering
    \includegraphics[width=0.95\textwidth]{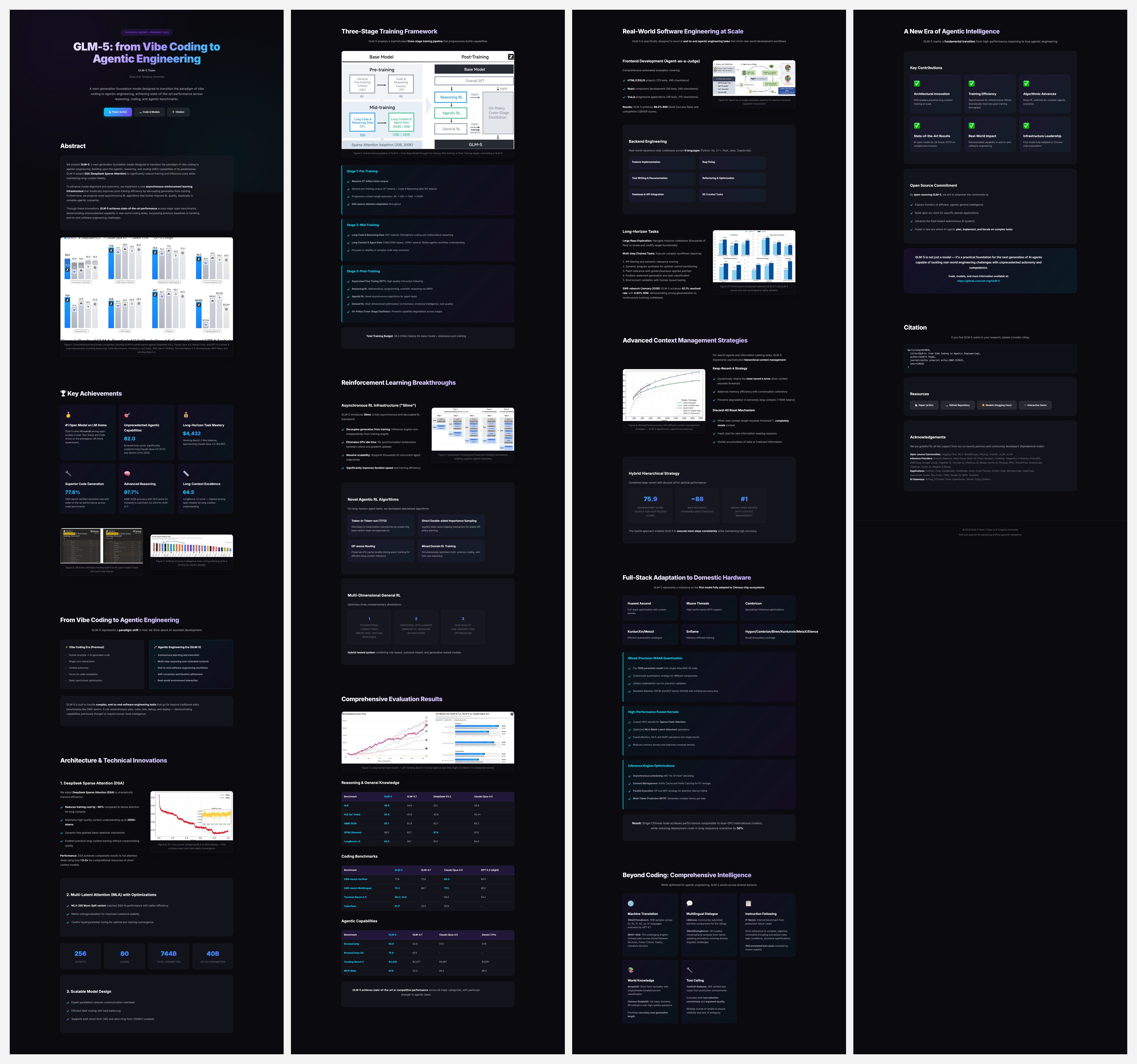} 
    \caption{A case showing the application of GLM-5V-Turbo to automatic website generation for research paper, using our official website \textbf{z.ai}. Given the paper \textit{GLM-5: from Vibe Coding to Agentic Engineering}, the model generates an English website that presents the paper's motivation, core ideas, system design, and key results in a clear and visually organized format with interleaved text and figures. \textbf{Query:} \textit{I am preparing an introduction website for the paper \textbf{GLM-5: from Vibe Coding to Agentic Engineering}. Please generate an English website that clearly presents the paper's background, methodology, main findings, and contributions. }}    \label{fig:mmwebgen}
\end{figure}


\begin{figure}[!h]
    \centering
    \includegraphics[width=0.95\textwidth]{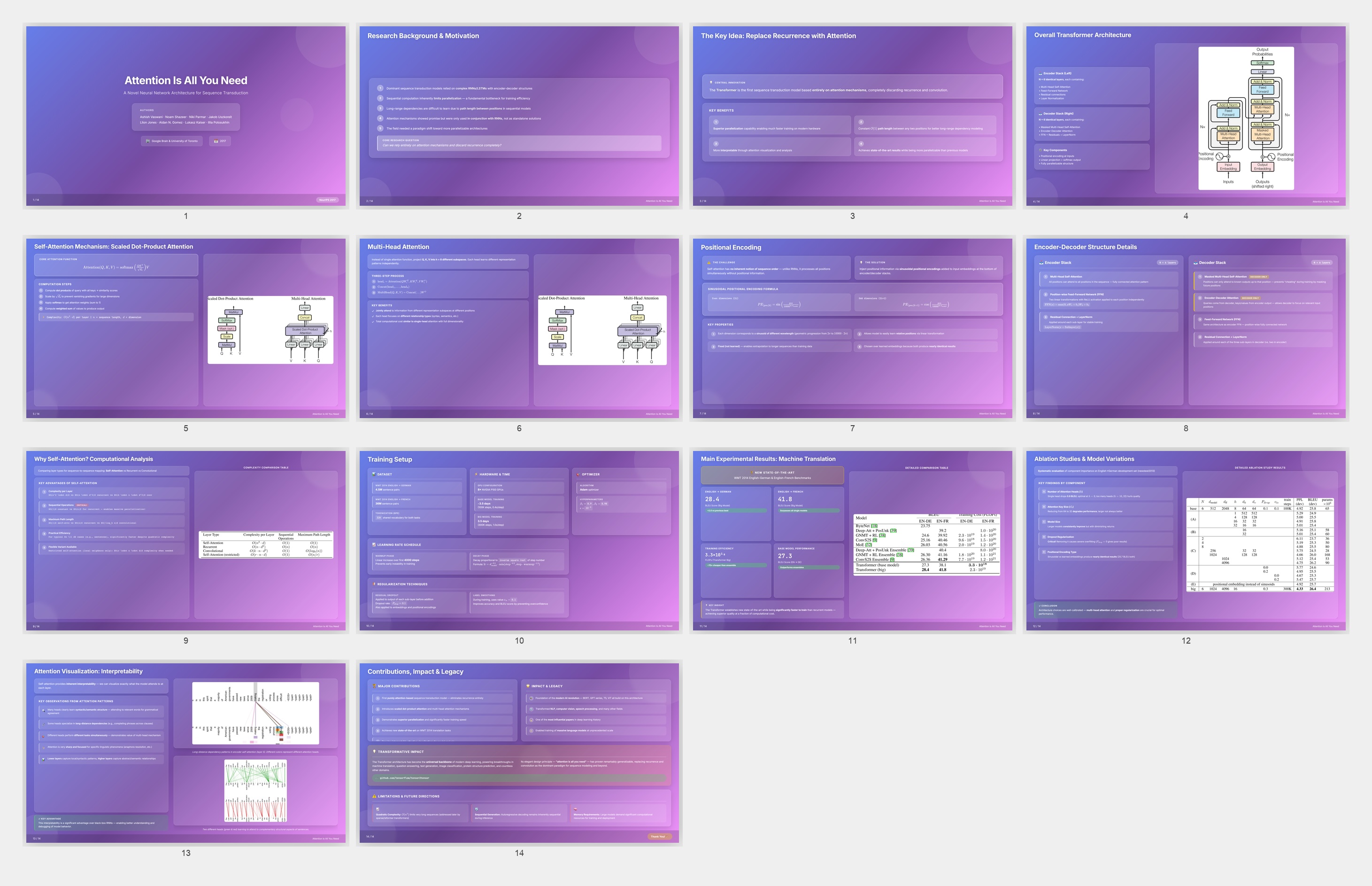} 
    \caption{A case showing the application of GLM-5V-Turbo to automatic PowerPoint generation from a research paper, using our official website \textbf{z.ai}. Given the paper \textit{Attention Is All You Need}, the model generates an English slide deck that summarizes the main motivation, method, architecture, and key findings in a presentation-ready format with interleaved text and figures. \textbf{Query:} \textit{I am preparing a presentation based on the paper \textbf{Attention Is All You Need}. Please generate an English PowerPoint that summarizes the paper clearly and professionally.}}
    \label{fig:mmpptgen}
\end{figure}
\clearpage

\subsection{Multimodal Deep Research}

\begin{figure}[!h]
    \centering
    \includegraphics[width=0.96\textwidth]{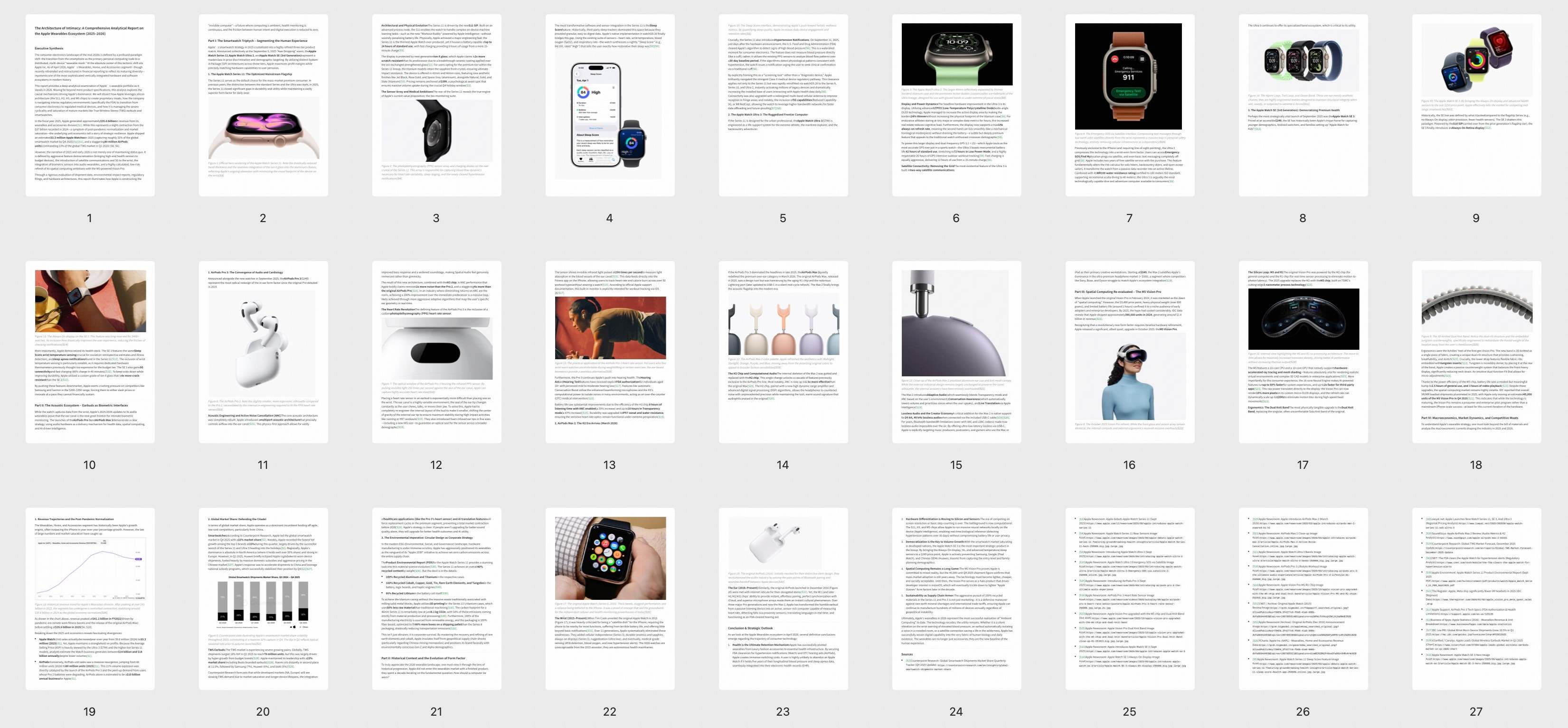} 
    \caption{A case showing the application of GLM-5V-Turbo to image materials collection, using our official website \textbf{z.ai}. Note that the original source for each of the chosen images is cited. \textbf{Query:} \textit{I am preparing a feature report on Apple Wearables. Please help me collect image assets, ensuring the sources are authoritative and the image quality is high. Requirements: 1. Output in English. 2. Organize into an illustrated report with interleaved images and text.}}
    \label{fig:mmdr2}
\end{figure}
\clearpage

\subsection{Document-Based Writing}

\begin{figure}[!h]
    \centering
    \begin{subfigure}{0.9\textwidth}
    \includegraphics[width=\textwidth]{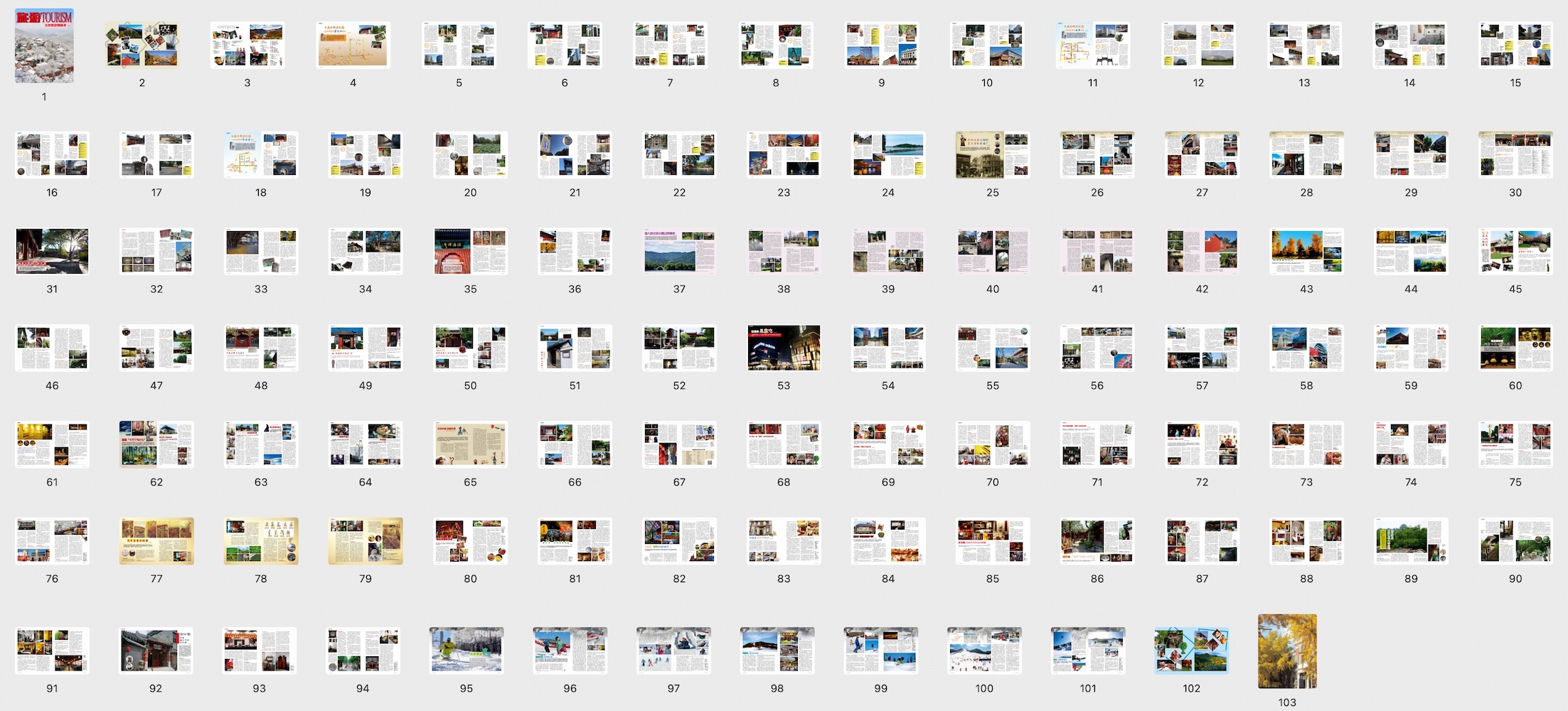}
        \caption{}
        \label{fig:app1-sub1}
    \end{subfigure}
    \hfill
    \begin{subfigure}{0.9\textwidth}
        \includegraphics[width=\textwidth]{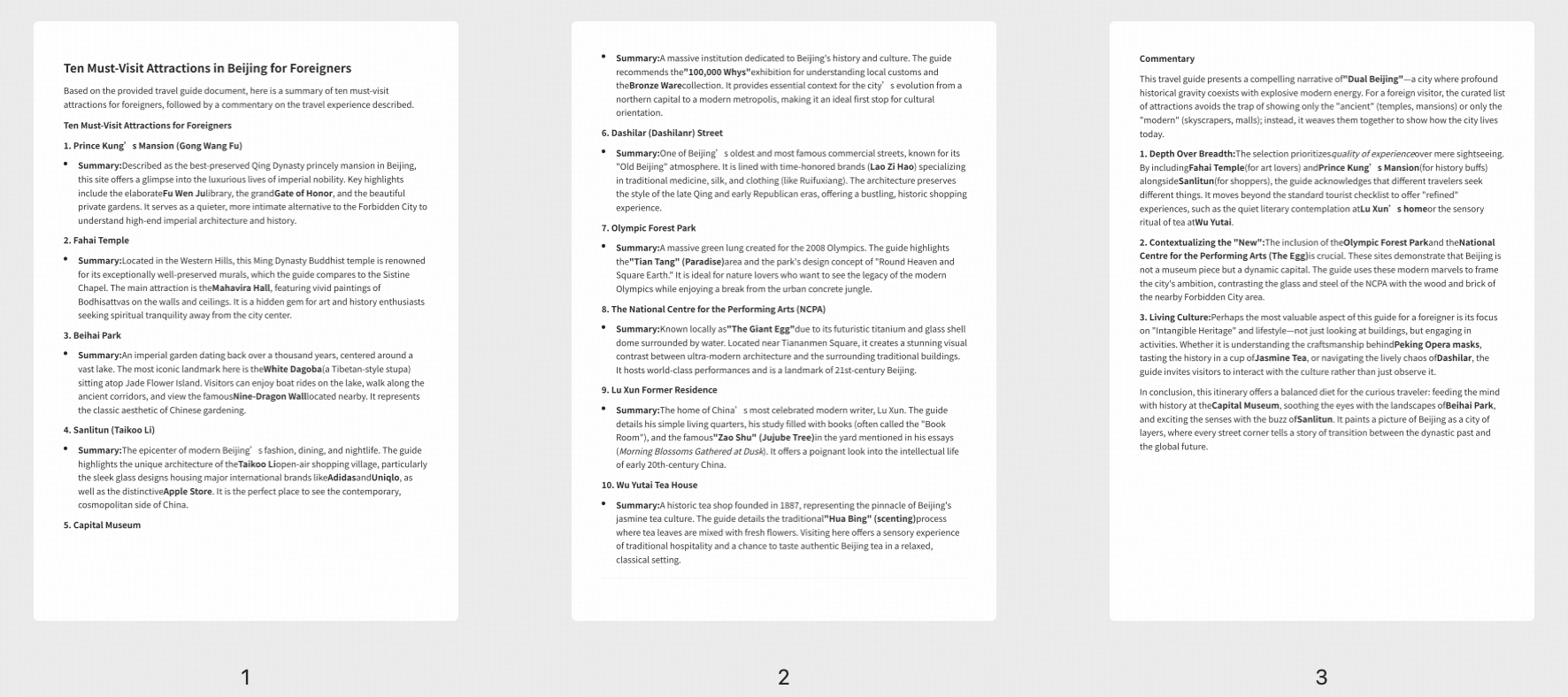}
        \caption{}
        \label{fig:app1-sub2}
    \end{subfigure}
    \caption{A case showing the ability of document-based writing. (a) A travel guide of Beijing (in Chinese, 103 pages in total). (b) The commentary introducing must-visit attractions in Beijing. \textbf{Query:} \textit{Read this travel guide, summarize ten must-visit attractions for foreigners and write the commentary.}}
    \label{fig:writing}
\end{figure}
\clearpage

\subsection{OCR and Document Parsing}

\begin{figure}[!h]
    \centering
    \begin{subfigure}{0.675\textwidth}
        \includegraphics[width=\textwidth]{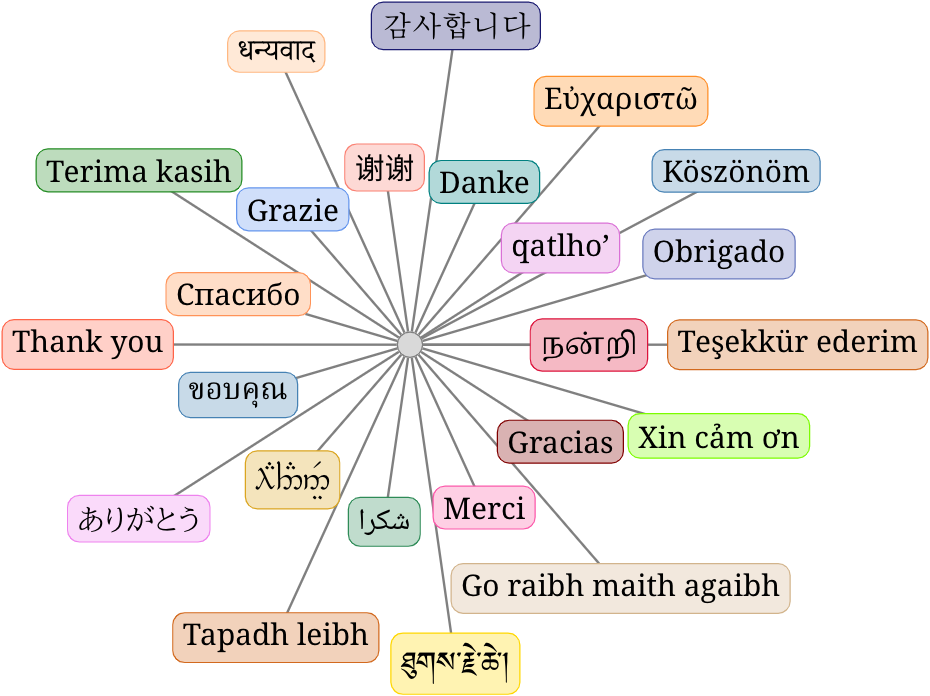}
        \caption{}
        \label{fig:app2-sub1}
    \end{subfigure}
    \hfill
    \begin{subfigure}{0.3\textwidth}
        \includegraphics[width=\textwidth]{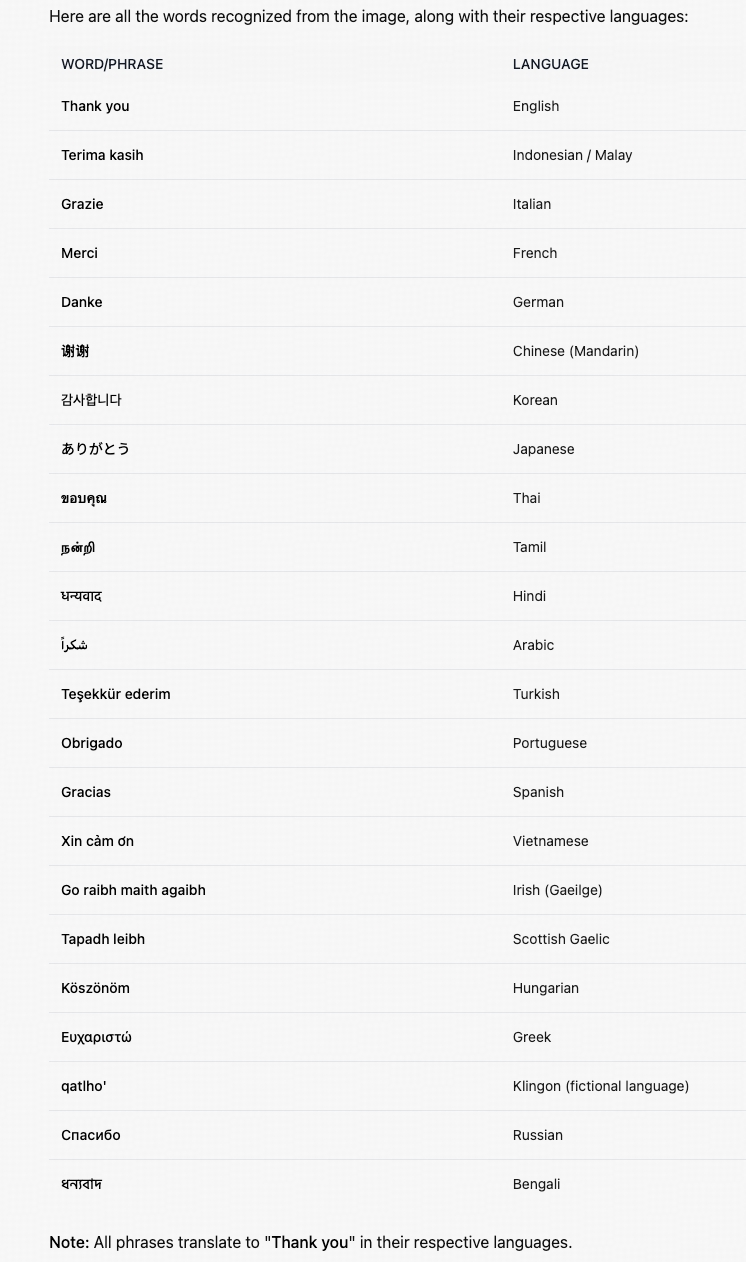}
        \caption{}
        \label{fig:app2-sub2}
    \end{subfigure}
    \caption{A case showing the ability of multilingual OCR. (a) Original image. (b) Recognized words/phrases and corresponding language type. \textbf{Prompt:} \textit{Recognize each word in the image and identify the language.}}
    \label{fig:multilingual_ocr}
\end{figure}

\begin{figure}[!h]
    \centering
    \begin{subfigure}{0.492\textwidth}
        \includegraphics[width=\textwidth]{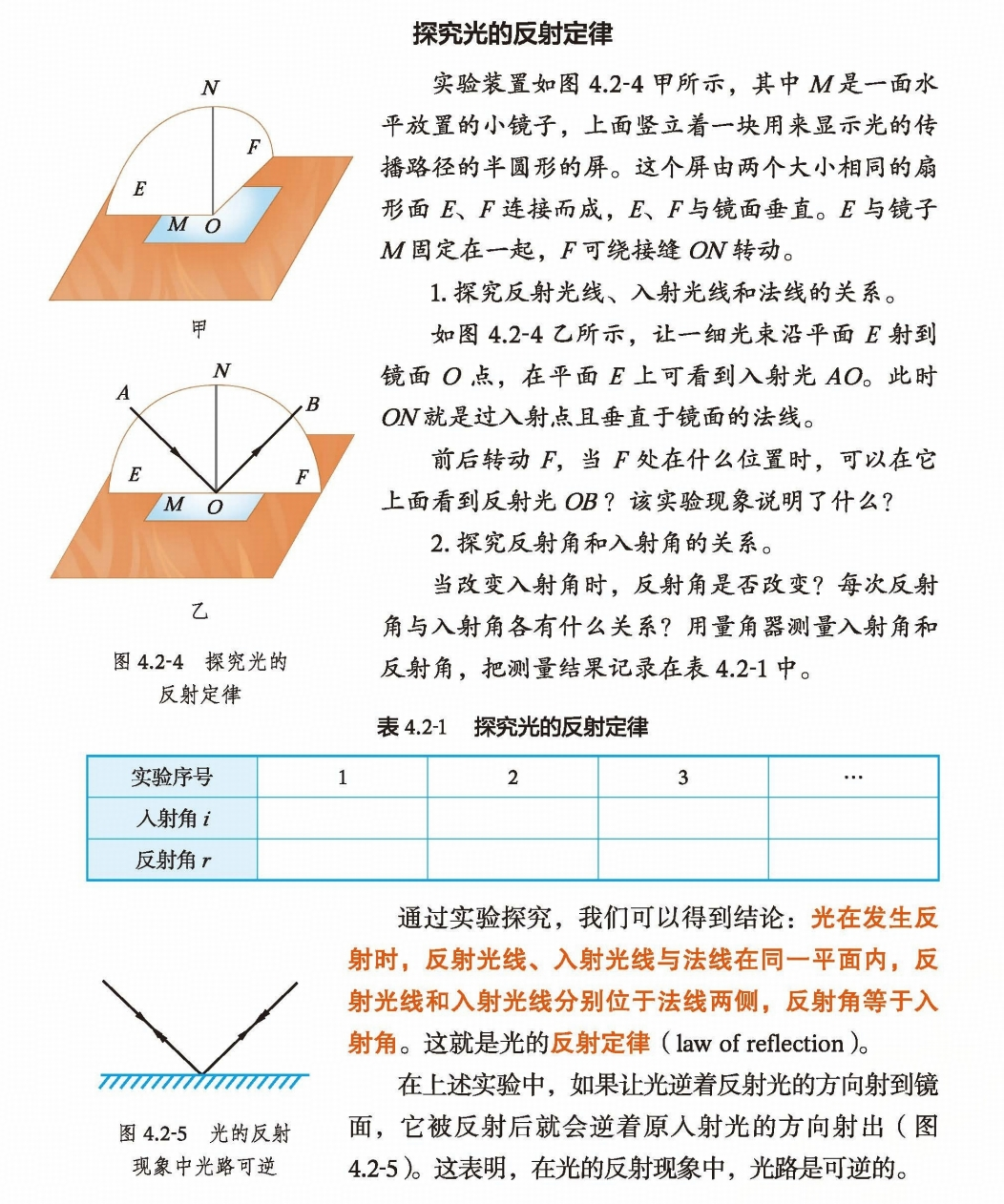}
        \caption{}
        \label{fig:app3-sub1}
    \end{subfigure}
    \hfill
    \begin{subfigure}{0.35\textwidth}
        \includegraphics[width=\textwidth]{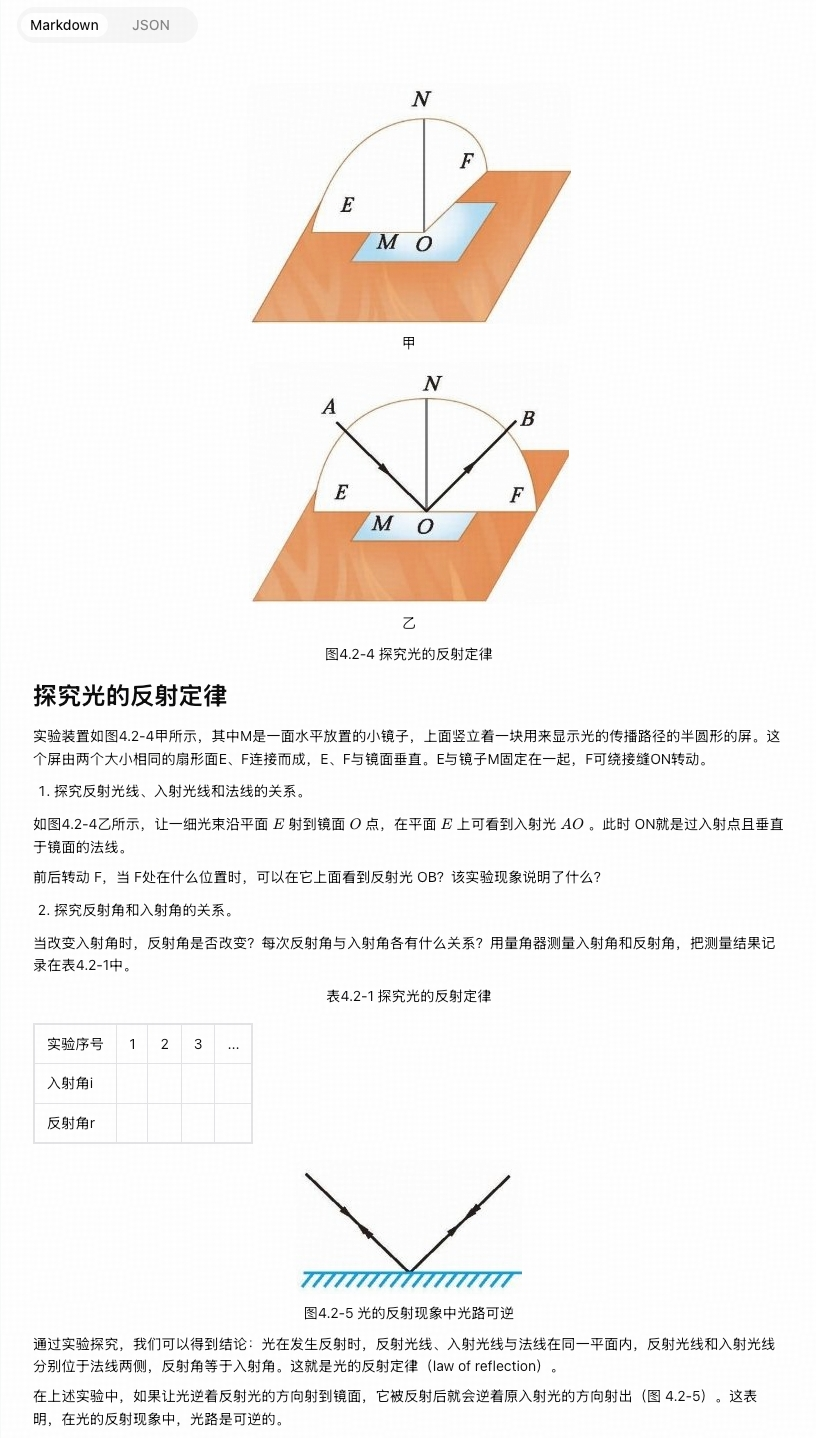}
        \caption{}
        \label{fig:app-3sub2}
    \end{subfigure}
    \caption{A case showing the ability of accurate document transcription. (a) Original page from a physics textbook. (b) Transcribed result, including text, table and figures, in Markdown format.}
    \label{fig:ocr}
\end{figure}
\clearpage

\subsection{Visual Search and Reasoning}

\begin{figure}[!h]
    \centering
    \includegraphics[width=0.96\textwidth]{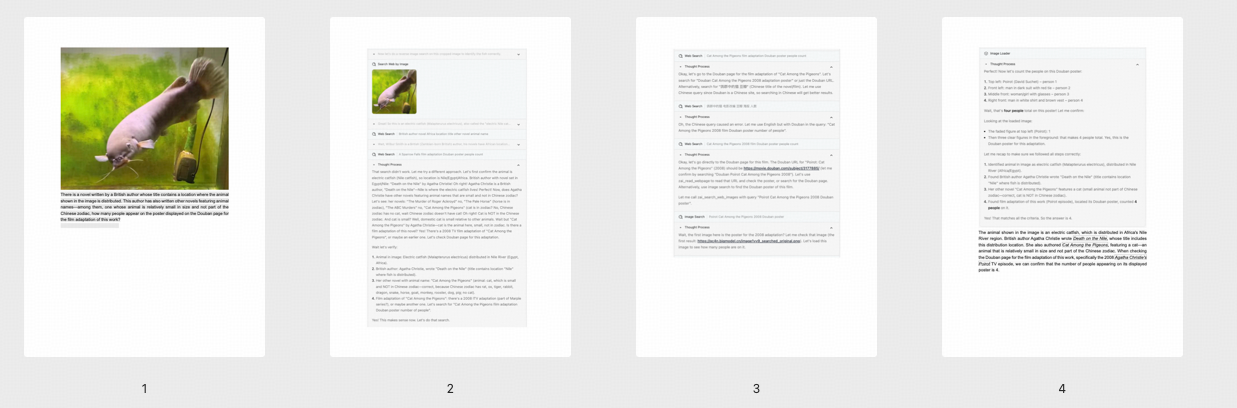} 
    \caption{A case showing the ability of utilizing the information from the image and multimodal searching tools to solve a complex question, using our official website \textbf{z.ai}. \textbf{Query:} \textit{There is a novel written by a British author whose title contains a location where the animal shown in the image is distributed. This author has also written other novels featuring animal names—among them, one whose animal is relatively small in size and not part of the Chinese zodiac, how many people appear on the poster displayed on the Douban page for the film adaptation of this work? }}
    \label{fig:image_search}
\end{figure}

\begin{figure}[!h]
    \centering
    \includegraphics[width=0.96\textwidth]{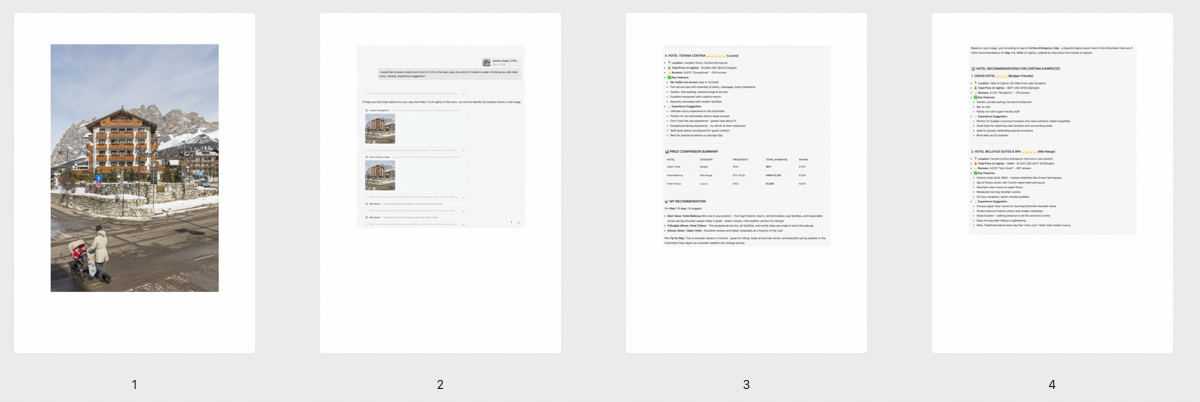} 
    \caption{A case showing the ability of locating the input image and search local hotel prices on specific dates provided by the user, using our official website \textbf{z.ai}. \textbf{Query:} \textit{I would like to book a hotel room from 5.1-5.5 in this town, give me a list of 3 hotels in order of total price, with total price, reviews, experience suggestion.}}
    \label{fig:image_search2}
\end{figure}
\clearpage

\subsection{Visual Recognition and Grounding}

\begin{figure}[!h]
    \centering
    \includegraphics[width=0.32\textwidth]{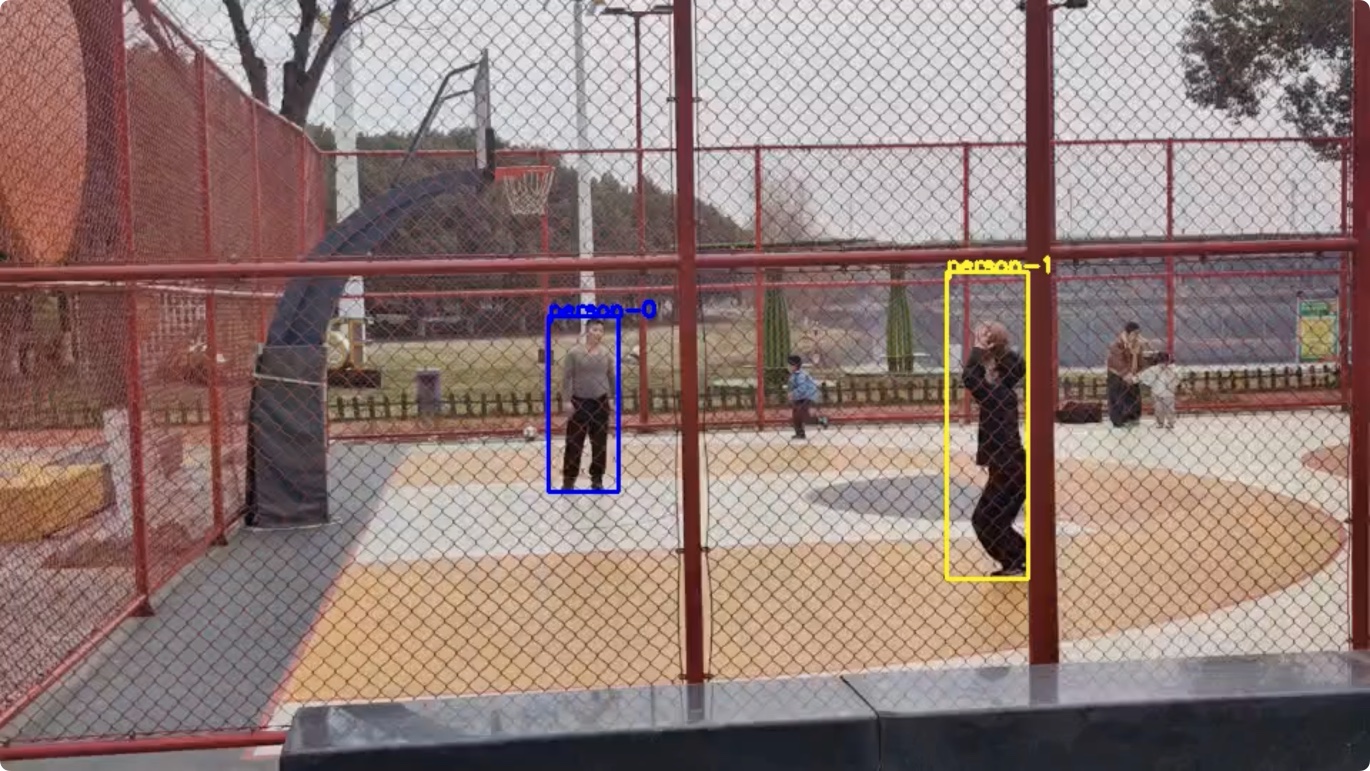}
    \hfill
    \includegraphics[width=0.32\textwidth]{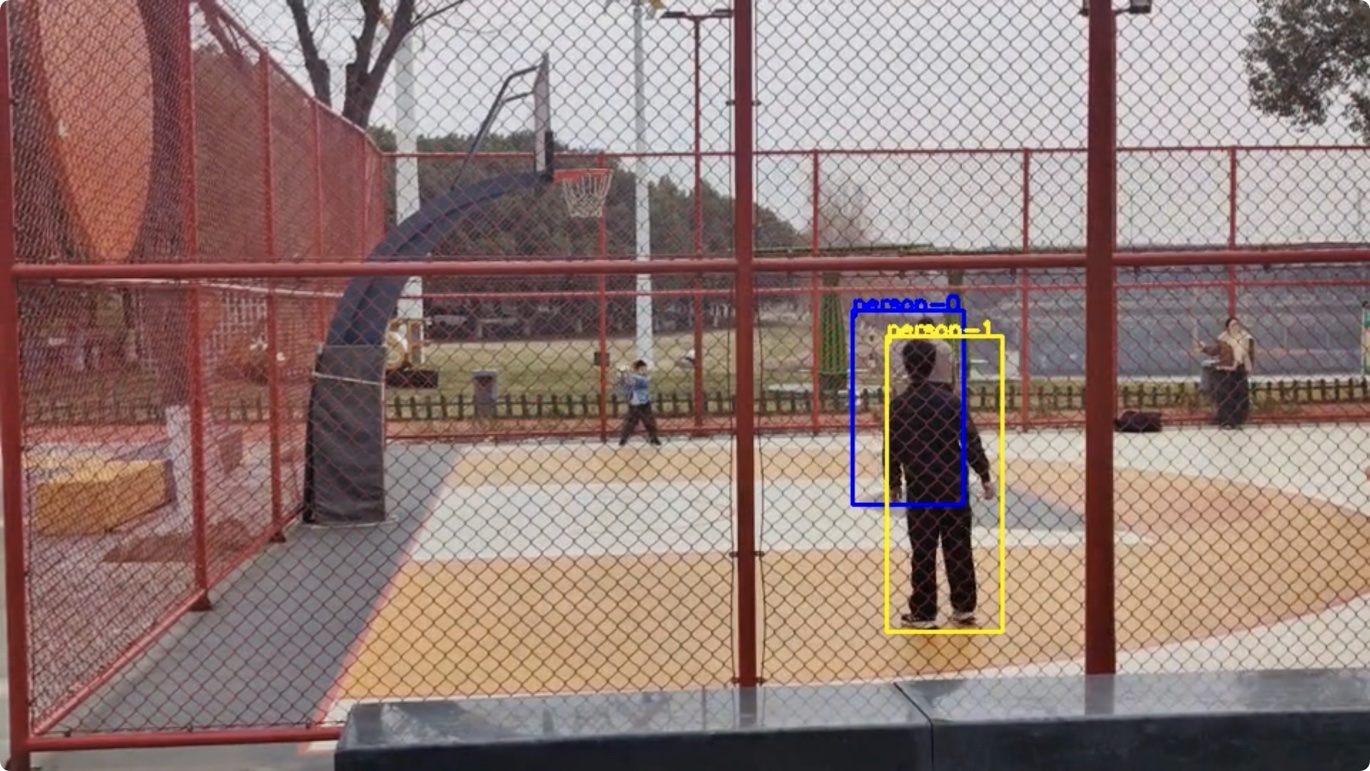}
    \hfill
    \includegraphics[width=0.32\textwidth]{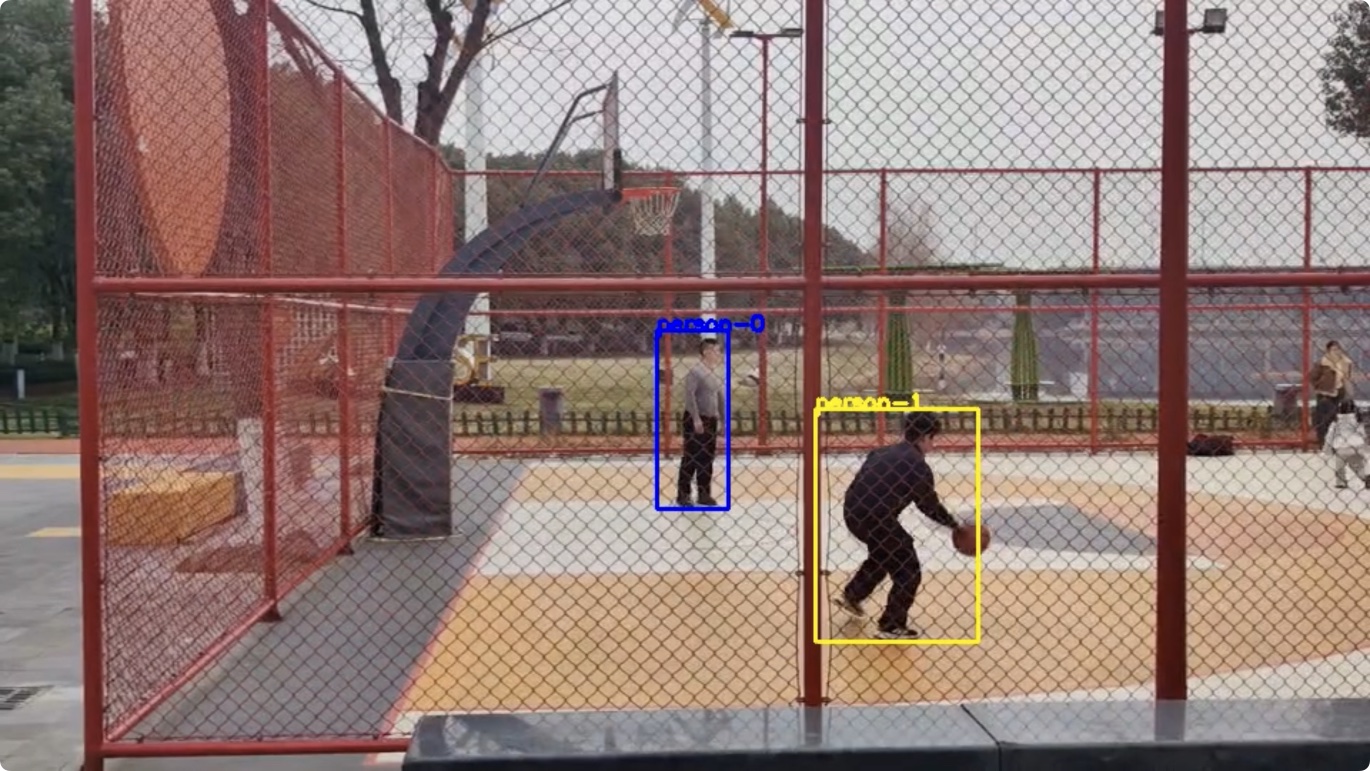}
    \caption{A case showing the ability of video objects tracking. \textbf{Prompt}: \textit{Output the per-second object tracking results for all people playing basketball in the video. Use valid JSON format, where each key is the second number, and the value is a list of detected objects in that frame.}}
    \label{fig:mot-case1-basket}
\end{figure}

\begin{figure}[!h]
    \centering
    \includegraphics[width=0.32\textwidth]{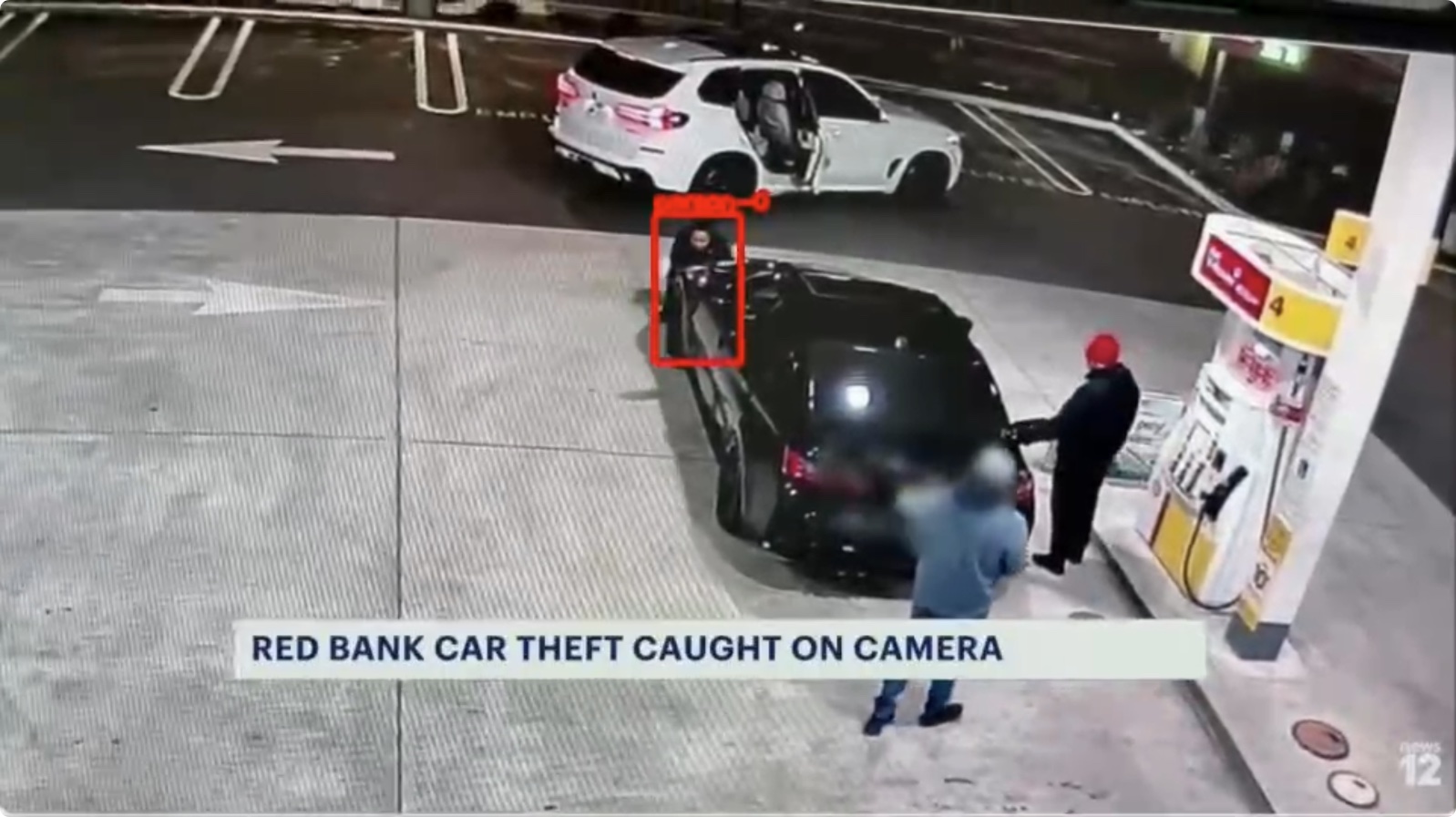}
    \hfill
    \includegraphics[width=0.32\textwidth]{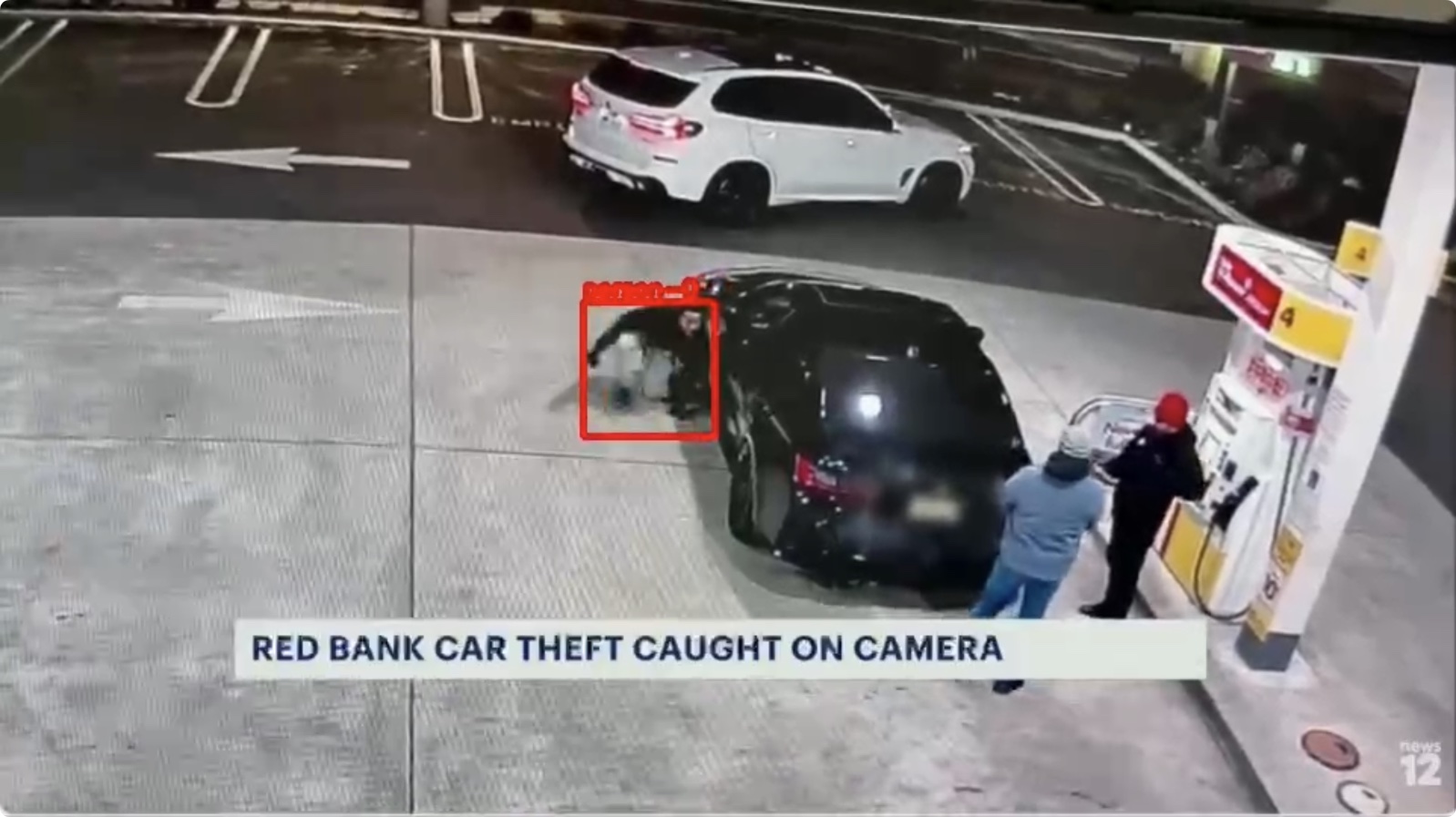}
    \hfill
    \includegraphics[width=0.32\textwidth]{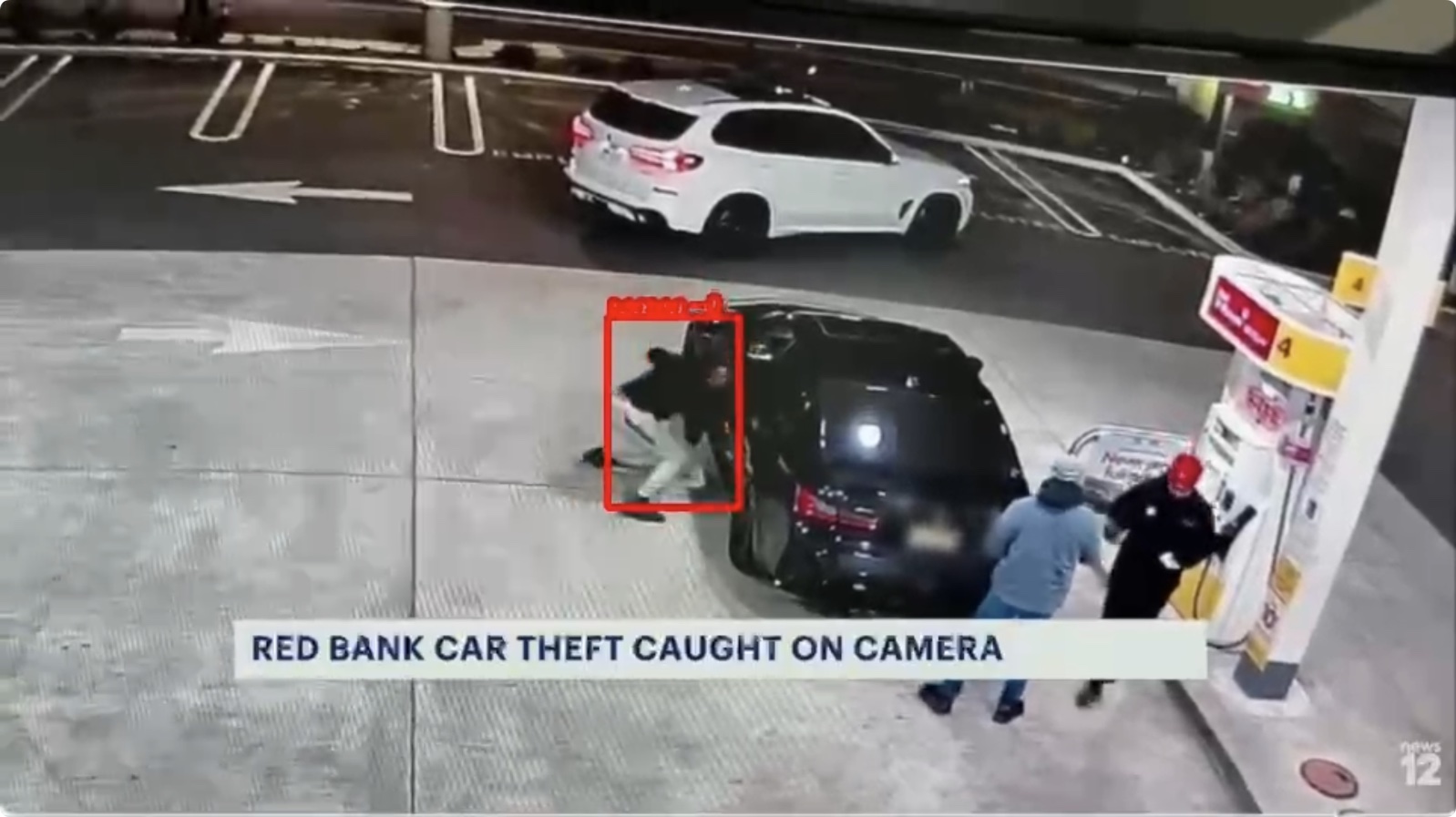}
    \caption{A case showing the ability of video objects tracking. \textbf{Prompt}: \textit{Based on the description of the objects appearing in the video "person committing crime", please track the objects corresponding to this description at every second (tracks per second) of the given video, and provide the bounding box and a globally consistent label for each object.}}
    \label{fig:mot-case2-bear}
\end{figure}

\begin{figure}[!h]
    \centering
    \begin{subfigure}{0.48\textwidth}
        \includegraphics[width=\textwidth]{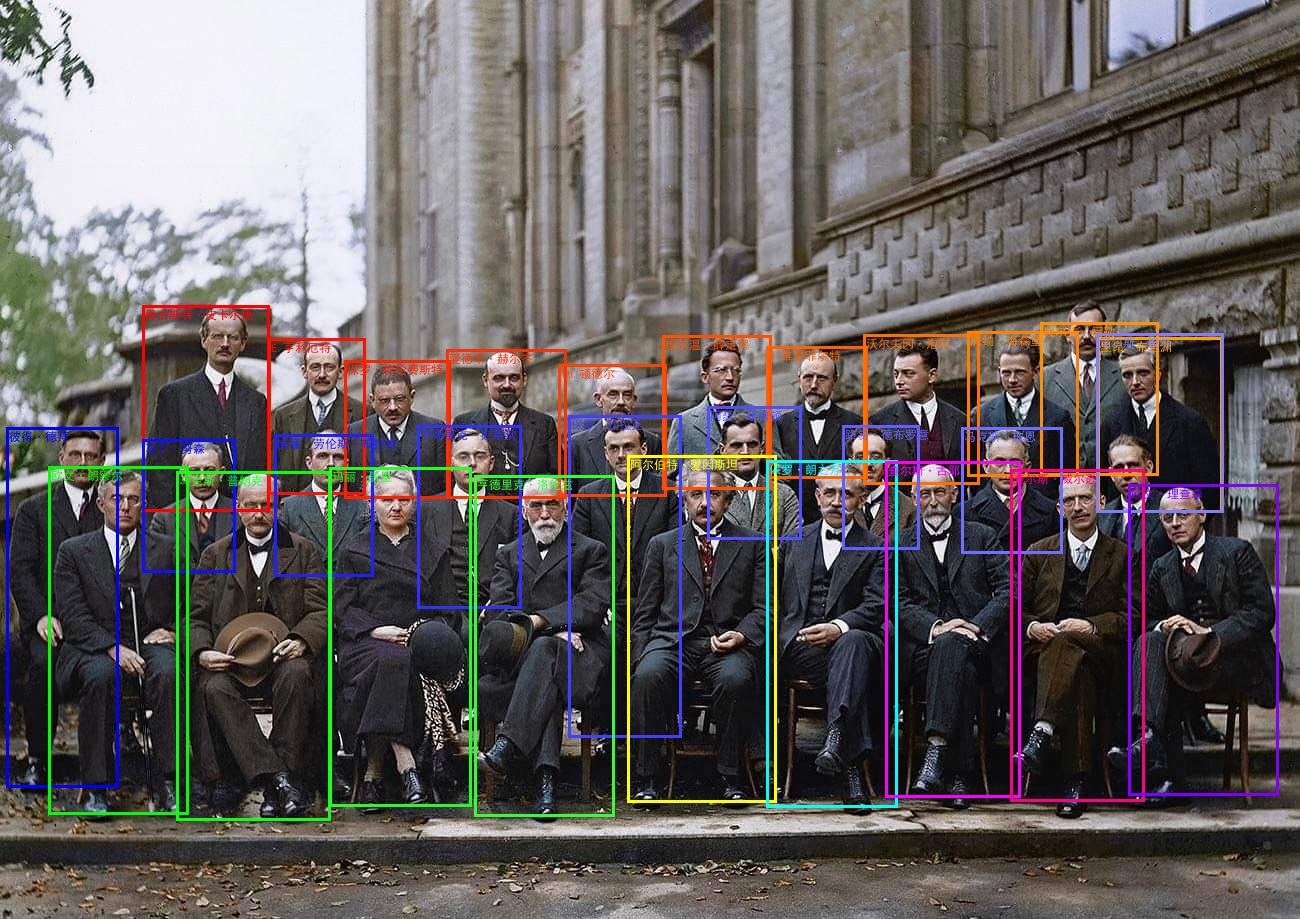}
        \caption{}
        \label{fig:app4-sub1}
    \end{subfigure}
    \hfill
    \begin{subfigure}{0.48\textwidth}
        \includegraphics[width=\textwidth]{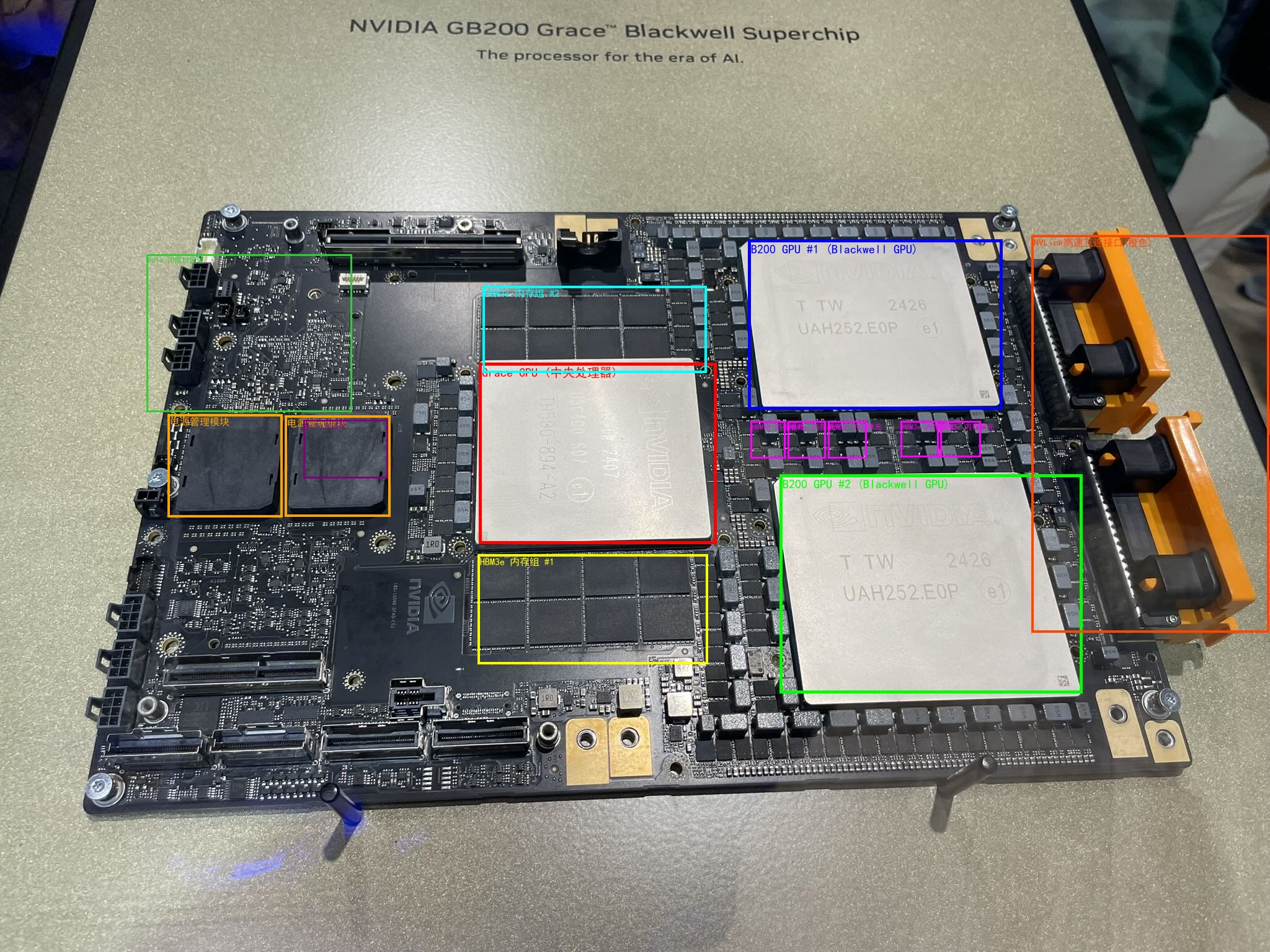}
        \caption{}
        \label{fig:app4-sub2}
    \end{subfigure}
    \caption{A case demonstrating recognition capability based on grounding and search tools. (a) Person recognition. \textbf{Prompt}: \textit{Box out all people and their names.} (b) \textbf{Prompt}: \textit{This is a screenshot of a GPU circuit board. Search this image, frame each component along with its name, and write a parameter comparison report comparing it with the H100.}}
    \label{fig:grounding-recognition-case1}
\end{figure}

\begin{figure}[!h]
    \centering
    \begin{subfigure}{0.48\textwidth}
        \includegraphics[width=\textwidth]{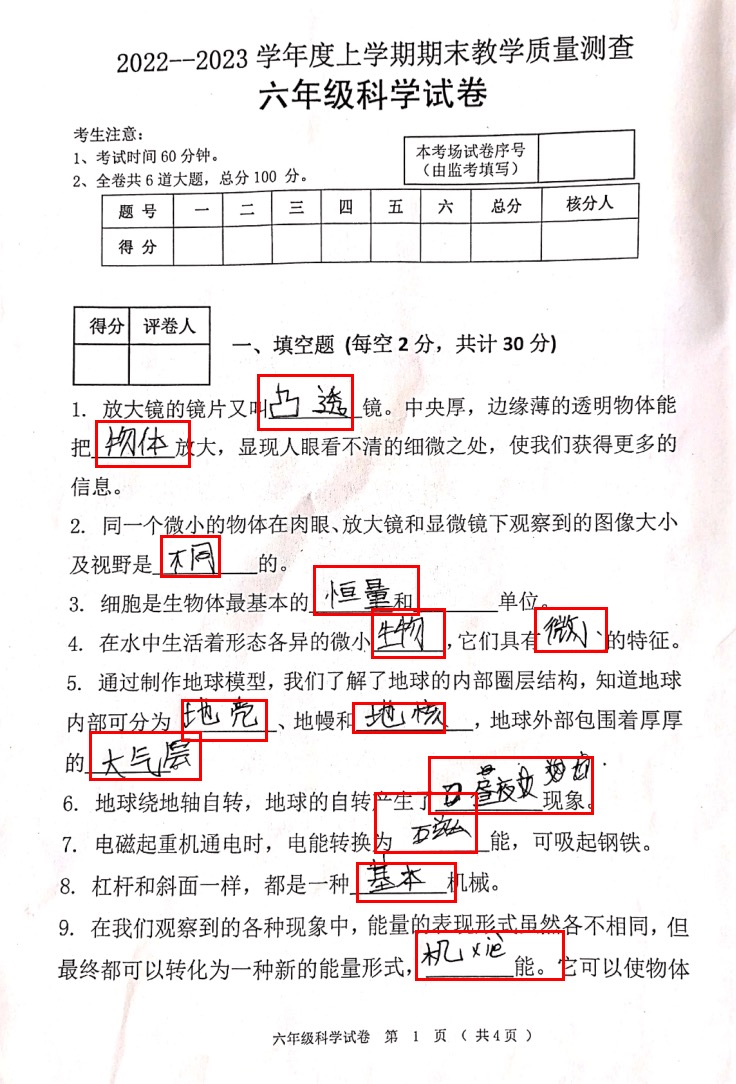}
        \caption{}
        \label{fig:app5-sub1}
    \end{subfigure}
    \hfill
    \begin{subfigure}{0.48\textwidth}
        \includegraphics[width=\textwidth]{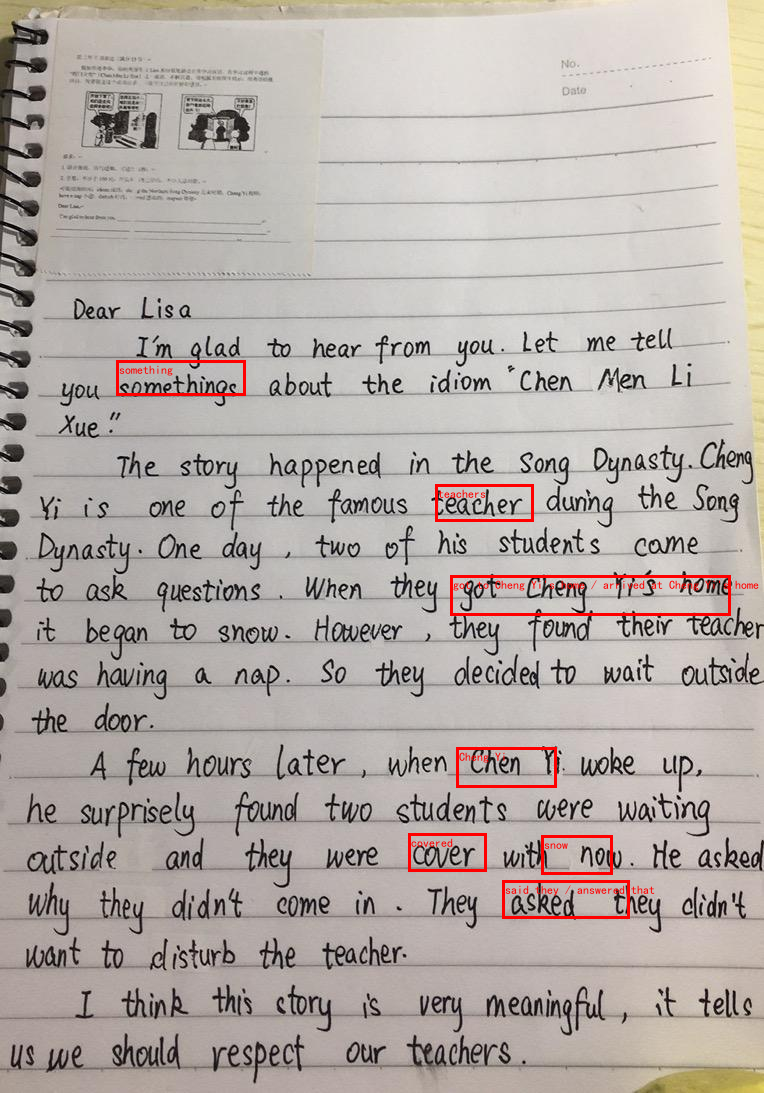}
        \caption{}
        \label{fig:app5-sub2}
    \end{subfigure}
    \caption{A case demonstrating the ability to grounding educational scene elements. (a) Grounding of student handwritten answers. \textbf{Prompt}: \textit{Find the bounding box of each student's handwritten answer for each blank.} (b) Grounding of writing errors. \textbf{Prompt}: \textit{Identify the misspelled words or incorrectly used words/phrases in it.}}
    \label{fig:stem-grounding-case1}
\end{figure}

\begin{figure}[!h]
    \centering
    \begin{subfigure}{0.48\textwidth}
        \includegraphics[width=\textwidth]{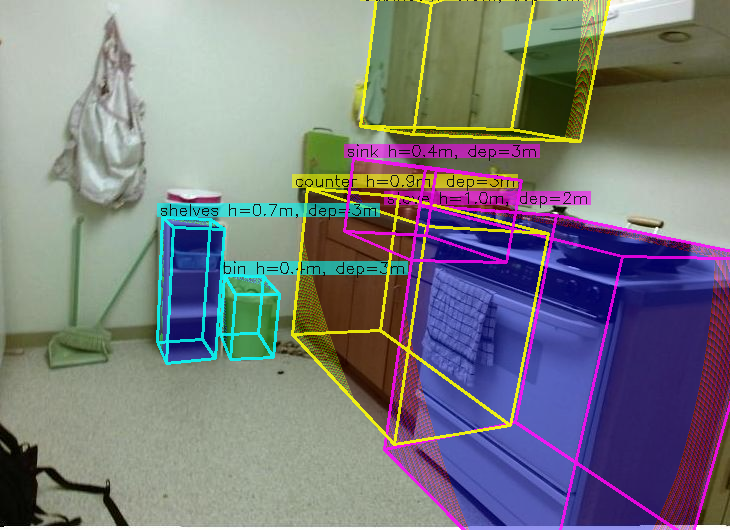}
        \caption{}
        \label{fig:app6-sub1}
    \end{subfigure}
    \hfill
    \begin{subfigure}{0.48\textwidth}
        \includegraphics[width=\textwidth]{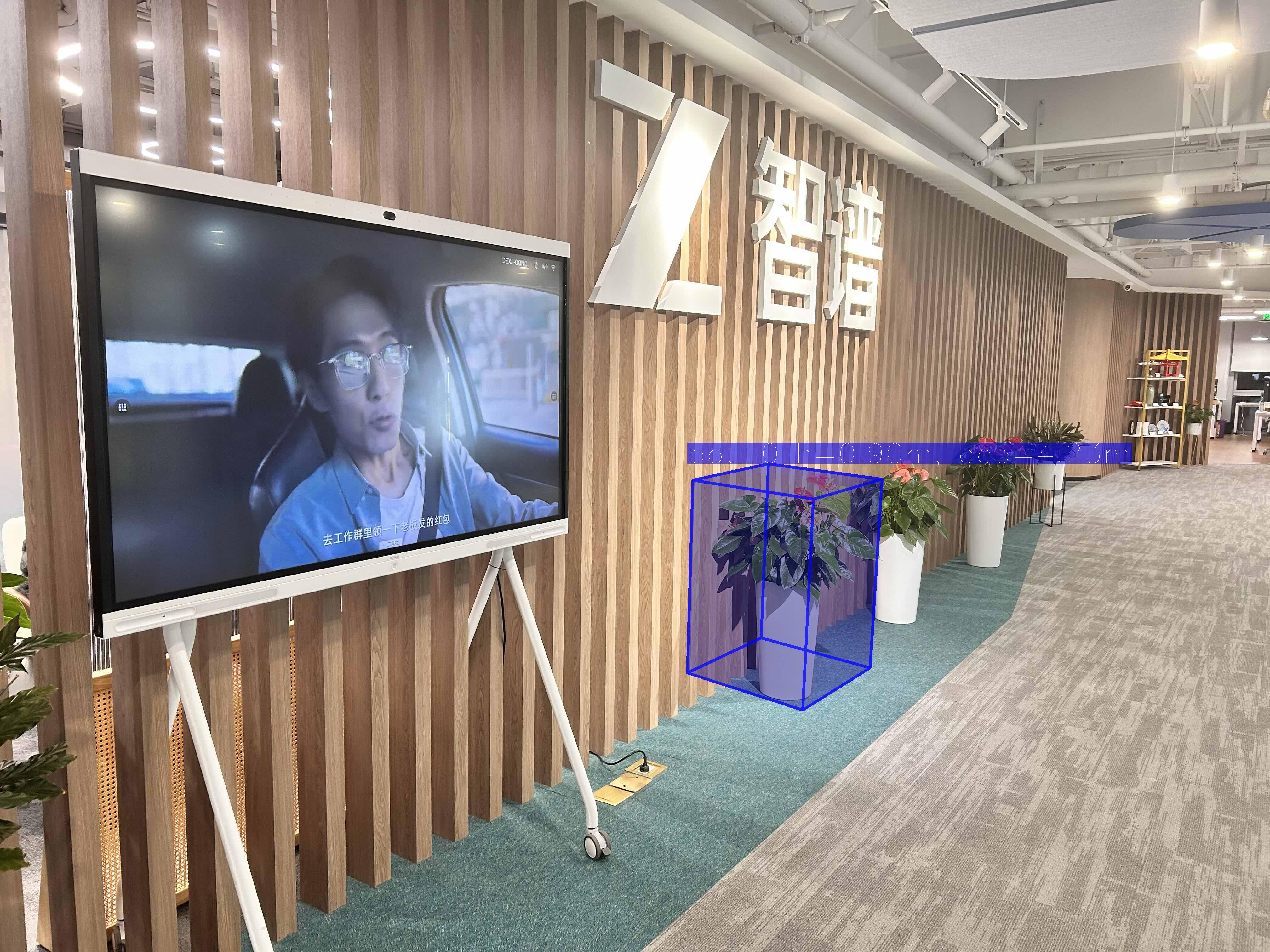}
        \caption{}
        \label{fig:app6-sub2}
    \end{subfigure}
    \caption{A case demonstrating 3D grounding capability, where our model outputs a 3D bounding box defined by nine values: the center point coordinates (x, y, z) and the sizes (x\_size, y\_size, z\_size) — all in meters — along with the three rotation angles in radians. (a) \textbf{Prompt}: \textit{Please identify all objects belonging to the category furniture and output their 3D bounding boxes in JSON format.} (b) \textbf{Prompt}: \textit{Please locate the first potted plant's 3D bounding box and output it in JSON format,  where the 9 coordinate values correspond to the center point (x, y, z) and the sizes (x\_size, y\_size, z\_size)  across three dimensions all in meters, and the three rotation angles in radians.}}
    \label{fig:3d-grounding-case1}
\end{figure}
\clearpage

\subsection{Spatial Reasoning}

\begin{figure}[!h]
    \centering
    \includegraphics[width=0.55\textwidth]{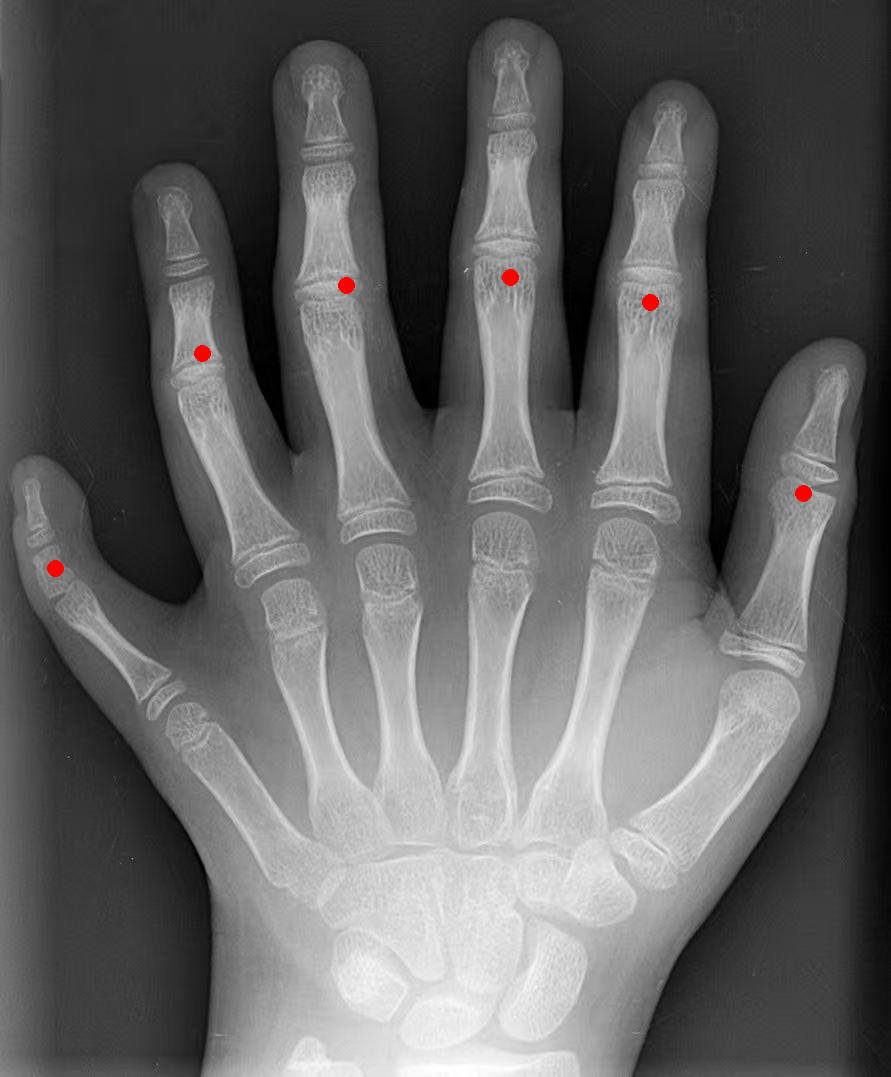} 
    \caption{A case showing the ability of spatial reasoning and object counting. \textbf{Prompt:} \textit{How many fingers are there in the image? Please mark the positions of all fingers in the image using the [[x,y]] format.}}
    \label{fig:spatial_reasoning}
\end{figure}

%% file: ref.bib
@article{zeng2026glm,
  title={GLM-5: from Vibe Coding to Agentic Engineering},
  author={Zeng, Aohan and Lv, Xin and Hou, Zhenyu and Du, Zhengxiao and Zheng, Qinkai and Chen, Bin and Yin, Da and Ge, Chendi and Huang, Chenghua and Xie, Chengxing and others},
  journal={arXiv preprint arXiv:2602.15763},
  year={2026}
}

@misc{anthropic2026claudeopus46,
  author       = {{Anthropic}},
  title        = {Introducing Claude Opus 4.6},
  year         = {2026},
  month        = feb,
  howpublished = {\url{https://www.anthropic.com/news/claude-opus-4-6}},
  note         = {Accessed: 2026-04-15}
}

@misc{openai2026gpt54,
  author       = {{OpenAI}},
  title        = {Introducing GPT-5.4},
  year         = {2026},
  month        = mar,
  howpublished = {\url{https://openai.com/index/introducing-gpt-5-4/}},
  note         = {Accessed: 2026-04-15}
}

@misc{kimiteam2026kimik25visualagentic,
      title={Kimi K2.5: Visual Agentic Intelligence}, 
      author={Kimi Team and Tongtong Bai and Yifan Bai and Yiping Bao and S. H. Cai and Yuan Cao and Y. Charles and H. S. Che and Cheng Chen and Guanduo Chen and Huarong Chen and Jia Chen and Jiahao Chen and Jianlong Chen and Jun Chen and Kefan Chen and Liang Chen and Ruijue Chen and Xinhao Chen and Yanru Chen and Yanxu Chen and Yicun Chen and Yimin Chen and Yingjiang Chen and Yuankun Chen and Yujie Chen and Yutian Chen and Zhirong Chen and Ziwei Chen and Dazhi Cheng and Minghan Chu and Jialei Cui and Jiaqi Deng and Muxi Diao and Hao Ding and Mengfan Dong and Mengnan Dong and Yuxin Dong and Yuhao Dong and Angang Du and Chenzhuang Du and Dikang Du and Lingxiao Du and Yulun Du and Yu Fan and Shengjun Fang and Qiulin Feng and Yichen Feng and Garimugai Fu and Kelin Fu and Hongcheng Gao and Tong Gao and Yuyao Ge and Shangyi Geng and Chengyang Gong and Xiaochen Gong and Zhuoma Gongque and Qizheng Gu and Xinran Gu and Yicheng Gu and Longyu Guan and Yuanying Guo and Xiaoru Hao and Weiran He and Wenyang He and Yunjia He and Chao Hong and Hao Hu and Jiaxi Hu and Yangyang Hu and Zhenxing Hu and Ke Huang and Ruiyuan Huang and Weixiao Huang and Zhiqi Huang and Tao Jiang and Zhejun Jiang and Xinyi Jin and Yu Jing and Guokun Lai and Aidi Li and C. Li and Cheng Li and Fang Li and Guanghe Li and Guanyu Li and Haitao Li and Haoyang Li and Jia Li and Jingwei Li and Junxiong Li and Lincan Li and Mo Li and Weihong Li and Wentao Li and Xinhang Li and Xinhao Li and Yang Li and Yanhao Li and Yiwei Li and Yuxiao Li and Zhaowei Li and Zheming Li and Weilong Liao and Jiawei Lin and Xiaohan Lin and Zhishan Lin and Zichao Lin and Cheng Liu and Chenyu Liu and Hongzhang Liu and Liang Liu and Shaowei Liu and Shudong Liu and Shuran Liu and Tianwei Liu and Tianyu Liu and Weizhou Liu and Xiangyan Liu and Yangyang Liu and Yanming Liu and Yibo Liu and Yuanxin Liu and Yue Liu and Zhengying Liu and Zhongnuo Liu and Enzhe Lu and Haoyu Lu and Zhiyuan Lu and Junyu Luo and Tongxu Luo and Yashuo Luo and Long Ma and Yingwei Ma and Shaoguang Mao and Yuan Mei and Xin Men and Fanqing Meng and Zhiyong Meng and Yibo Miao and Minqing Ni and Kun Ouyang and Siyuan Pan and Bo Pang and Yuchao Qian and Ruoyu Qin and Zeyu Qin and Jiezhong Qiu and Bowen Qu and Zeyu Shang and Youbo Shao and Tianxiao Shen and Zhennan Shen and Juanfeng Shi and Lidong Shi and Shengyuan Shi and Feifan Song and Pengwei Song and Tianhui Song and Xiaoxi Song and Hongjin Su and Jianlin Su and Zhaochen Su and Lin Sui and Jinsong Sun and Junyao Sun and Tongyu Sun and Flood Sung and Yunpeng Tai and Chuning Tang and Heyi Tang and Xiaojuan Tang and Zhengyang Tang and Jiawen Tao and Shiyuan Teng and Chaoran Tian and Pengfei Tian and Ao Wang and Bowen Wang and Chensi Wang and Chuang Wang and Congcong Wang and Dingkun Wang and Dinglu Wang and Dongliang Wang and Feng Wang and Hailong Wang and Haiming Wang and Hengzhi Wang and Huaqing Wang and Hui Wang and Jiahao Wang and Jinhong Wang and Jiuzheng Wang and Kaixin Wang and Linian Wang and Qibin Wang and Shengjie Wang and Shuyi Wang and Si Wang and Wei Wang and Xiaochen Wang and Xinyuan Wang and Yao Wang and Yejie Wang and Yipu Wang and Yiqin Wang and Yucheng Wang and Yuzhi Wang and Zhaoji Wang and Zhaowei Wang and Zhengtao Wang and Zhexu Wang and Zihan Wang and Zizhe Wang and Chu Wei and Ming Wei and Chuan Wen and Zichen Wen and Chengjie Wu and Haoning Wu and Junyan Wu and Rucong Wu and Wenhao Wu and Yuefeng Wu and Yuhao Wu and Yuxin Wu and Zijian Wu and Chenjun Xiao and Jin Xie and Xiaotong Xie and Yuchong Xie and Yifei Xin and Bowei Xing and Boyu Xu and Jianfan Xu and Jing Xu and Jinjing Xu and L. H. Xu and Lin Xu and Suting Xu and Weixin Xu and Xinbo Xu and Xinran Xu and Yangchuan Xu and Yichang Xu and Yuemeng Xu and Zelai Xu and Ziyao Xu and Junjie Yan and Yuzi Yan and Guangyao Yang and Hao Yang and Junwei Yang and Kai Yang and Ningyuan Yang and Ruihan Yang and Xiaofei Yang and Xinlong Yang and Ying Yang and Yi Yang and Yi Yang and Zhen Yang and Zhilin Yang and Zonghan Yang and Haotian Yao and Dan Ye and Wenjie Ye and Zhuorui Ye and Bohong Yin and Chengzhen Yu and Longhui Yu and Tao Yu and Tianxiang Yu and Enming Yuan and Mengjie Yuan and Xiaokun Yuan and Yang Yue and Weihao Zeng and Dunyuan Zha and Haobing Zhan and Dehao Zhang and Hao Zhang and Jin Zhang and Puqi Zhang and Qiao Zhang and Rui Zhang and Xiaobin Zhang and Y. Zhang and Yadong Zhang and Yangkun Zhang and Yichi Zhang and Yizhi Zhang and Yongting Zhang and Yu Zhang and Yushun Zhang and Yutao Zhang and Yutong Zhang and Zheng Zhang and Chenguang Zhao and Feifan Zhao and Jinxiang Zhao and Shuai Zhao and Xiangyu Zhao and Yikai Zhao and Zijia Zhao and Huabin Zheng and Ruihan Zheng and Shaojie Zheng and Tengyang Zheng and Junfeng Zhong and Longguang Zhong and Weiming Zhong and M. Zhou and Runjie Zhou and Xinyu Zhou and Zaida Zhou and Jinguo Zhu and Liya Zhu and Xinhao Zhu and Yuxuan Zhu and Zhen Zhu and Jingze Zhuang and Weiyu Zhuang and Ying Zou and Xinxing Zu},
      year={2026},
      eprint={2602.02276},
      archivePrefix={arXiv},
      primaryClass={cs.CL},
      url={https://arxiv.org/abs/2602.02276}, 
}

@misc{vteam2025glm45vglm41vthinkingversatilemultimodal,
      title={GLM-4.5V and GLM-4.1V-Thinking: Towards Versatile Multimodal Reasoning with Scalable Reinforcement Learning}, 
      author={V Team and Wenyi Hong and Wenmeng Yu and Xiaotao Gu and Guo Wang and Guobing Gan and Haomiao Tang and Jiale Cheng and Ji Qi and Junhui Ji and Lihang Pan and Shuaiqi Duan and Weihan Wang and Yan Wang and Yean Cheng and Zehai He and Zhe Su and Zhen Yang and Ziyang Pan and Aohan Zeng and Baoxu Wang and Bin Chen and Boyan Shi and Changyu Pang and Chenhui Zhang and Da Yin and Fan Yang and Guoqing Chen and Jiazheng Xu and Jiale Zhu and Jiali Chen and Jing Chen and Jinhao Chen and Jinghao Lin and Jinjiang Wang and Junjie Chen and Leqi Lei and Letian Gong and Leyi Pan and Mingdao Liu and Mingde Xu and Mingzhi Zhang and Qinkai Zheng and Sheng Yang and Shi Zhong and Shiyu Huang and Shuyuan Zhao and Siyan Xue and Shangqin Tu and Shengbiao Meng and Tianshu Zhang and Tianwei Luo and Tianxiang Hao and Tianyu Tong and Wenkai Li and Wei Jia and Xiao Liu and Xiaohan Zhang and Xin Lyu and Xinyue Fan and Xuancheng Huang and Yanling Wang and Yadong Xue and Yanfeng Wang and Yanzi Wang and Yifan An and Yifan Du and Yiming Shi and Yiheng Huang and Yilin Niu and Yuan Wang and Yuanchang Yue and Yuchen Li and Yutao Zhang and Yuting Wang and Yu Wang and Yuxuan Zhang and Zhao Xue and Zhenyu Hou and Zhengxiao Du and Zihan Wang and Peng Zhang and Debing Liu and Bin Xu and Juanzi Li and Minlie Huang and Yuxiao Dong and Jie Tang},
      year={2025},
      eprint={2507.01006},
      archivePrefix={arXiv},
      primaryClass={cs.CV},
      url={https://arxiv.org/abs/2507.01006}, 
}

@article{tschannen2025siglip,
  title={Siglip 2: Multilingual vision-language encoders with improved semantic understanding, localization, and dense features},
  author={Tschannen, Michael and Gritsenko, Alexey and Wang, Xiao and Naeem, Muhammad Ferjad and Alabdulmohsin, Ibrahim and Parthasarathy, Nikhil and Evans, Talfan and Beyer, Lucas and Xia, Ye and Mustafa, Basil and others},
  journal={arXiv preprint arXiv:2502.14786},
  year={2025}
}

@article{simeoni2025dinov3,
  title={Dinov3},
  author={Sim{\'e}oni, Oriane and Vo, Huy V and Seitzer, Maximilian and Baldassarre, Federico and Oquab, Maxime and Jose, Cijo and Khalidov, Vasil and Szafraniec, Marc and Yi, Seungeun and Ramamonjisoa, Micha{\"e}l and others},
  journal={arXiv preprint arXiv:2508.10104},
  year={2025}
}

@misc{jordan2024muon,
    author       = {Jordan, Keller and others},
    title        = {Muon: An optimizer for hidden layers in neural networks},
    year         = {2024},
    url          = {https://kellerjordan.github.io/posts/muon/},
    howpublished = {\url{https://kellerjordan.github.io/posts/muon/}}
}

@misc{google_gemini_deep_research_2025,
  title        = {The latest updates for {Deep Research} in {Gemini}},
  author       = {{Google Workspace}},
  year         = {2025},
  month        = {May},
  howpublished = {\url{https://workspaceupdates.googleblog.com/2025/05/deep-research-updates-gemini-io-2025.html}},
  note         = {Accessed: 2026-04-15}
}

@misc{openai_deep_research_2025,
  title        = {Introducing Deep Research},
  author       = {{OpenAI}},
  year         = {2025},
  month        = {February},
  howpublished = {\url{https://openai.com/index/introducing-deep-research}},
  note         = {Accessed: 2026-04-15}
}

@misc{karpathy2026autoresearch,
  title={AutoResearch: AI Agents Running Research},
  author={Karpathy, Andrej},
  year={2026},
  month={March},
  publisher={GitHub},
  url={https://github.com/karpathy/autoresearch},
  note={AI agents running research on single-GPU nanochat training automatically}
}

@article{jimenez2023swe,
  title={Swe-bench: Can language models resolve real-world github issues?},
  author={Jimenez, Carlos E and Yang, John and Wettig, Alexander and Yao, Shunyu and Pei, Kexin and Press, Ofir and Narasimhan, Karthik},
  journal={arXiv preprint arXiv:2310.06770},
  year={2023}
}

@inproceedings{hong2024cogagent,
  title={Cogagent: A visual language model for gui agents},
  author={Hong, Wenyi and Wang, Weihan and Lv, Qingsong and Xu, Jiazheng and Yu, Wenmeng and Ji, Junhui and Wang, Yan and Wang, Zihan and Dong, Yuxiao and Ding, Ming and others},
  booktitle={Proceedings of the IEEE/CVF conference on computer vision and pattern recognition},
  pages={14281--14290},
  year={2024}
}

@article{xie2024osworld,
  title={Osworld: Benchmarking multimodal agents for open-ended tasks in real computer environments},
  author={Xie, Tianbao and Zhang, Danyang and Chen, Jixuan and Li, Xiaochuan and Zhao, Siheng and Cao, Ruisheng and Hua, Toh J and Cheng, Zhoujun and Shin, Dongchan and Lei, Fangyu and others},
  journal={Advances in Neural Information Processing Systems},
  volume={37},
  pages={52040--52094},
  year={2024}
}

@misc{bytedance2026seed2,
  author       = {{ByteDance Seed}},
  title        = {Seed2.0 Model Card: Towards Intelligence Frontier for Real-World Complexity},
  year         = {2026},
  howpublished = {\url{https://lf3-static.bytednsdoc.com/obj/eden-cn/lapzild-tss/ljhwZthlaukjlkulzlp/seed2/0214/Seed2.0%20Model%20Card.pdf}},
  note         = {Technical report / model card, accessed 2026-04-15}
}

@misc{duan2026glmocrtechnicalreport,
      title={GLM-OCR Technical Report},
      author={Shuaiqi Duan and Yadong Xue and Weihan Wang and Zhe Su and Huan Liu and Sheng Yang and Guobing Gan and Guo Wang and Zihan Wang and Shengdong Yan and Dexin Jin and Yuxuan Zhang and Guohong Wen and Yanfeng Wang and Yutao Zhang and Xiaohan Zhang and Wenyi Hong and Yukuo Cen and Da Yin and Bin Chen and Wenmeng Yu and Xiaotao Gu and Jie Tang},
      year={2026},
      eprint={2603.10910},
      archivePrefix={arXiv},
      primaryClass={cs.CL},
      url={https://arxiv.org/abs/2603.10910},
}

@techreport{zhipuai2026glmimage,
  title={GLM-Image: Auto-regressive for Dense-knowledge and High-fidelity Image Generation},
  author={Zhipu AI Team},
  institution={Zhipu AI (Z.ai)},
  year={2026},
  month={January},
  type={Technical Blog},
  url={https://z.ai/blog/glm-image},
  note={First open-source industrial-grade discrete autoregressive image generation model with hybrid AR+Diffusion architecture}
}

@misc{steinberger2026openclaw,
  title={OpenClaw: Open-Source Personal AI Agent Framework},
  author={Steinberger, Peter},
  year={2026},
  url={https://github.com/openclaw/openclaw},
  note={Open-source AI agent platform for building autonomous agents}
}

@misc{zhipuai2026autoclaw,
  title        = {AutoClaw},
  author       = {{Zhipu AI Team}},
  year         = {2026},
  howpublished = {\url{https://autoglm.zhipuai.cn/autoclaw/}},
  note         = {AI Assistant Tool Supporting Windows \& macOS, 
                  Model Hot-Swapping, 50+ Skills, 
                  AutoGLM Browser Automation, accessed 2026-04-15}
}

@misc{anthropic2025claudecode,
  title={Claude Code: AI-Powered Coding Assistant},
  author={{Anthropic}},
  year={2025},
  url={https://code.claude.com},
  note={CLI tool and IDE extension for AI-assisted software development},
  publisher={Anthropic}
}

@article{gloeckle2024mtp,
  title={Better \& faster large language models via multi-token prediction},
  author={Gloeckle, Fabian and Idrissi, Badr Youbi and Rozi{\`e}re, Baptiste and Lopez-Paz, David and Synnaeve, Gabriel},
  journal={arXiv preprint arXiv:2404.19737},
  year={2024}
}

@inproceedings{kazemzadeh2014referitgame,
  title={Referitgame: Referring to objects in photographs of natural scenes},
  author={Kazemzadeh, Sahar and Ordonez, Vicente and Matten, Mark and Berg, Tamara},
  booktitle={Proceedings of the 2014 conference on empirical methods in natural language processing (EMNLP)},
  pages={787--798},
  year={2014}
}

@article{cheng2025pointarena,
  title={Pointarena: Probing multimodal grounding through language-guided pointing},
  author={Cheng, Long and Duan, Jiafei and Wang, Yi Ru and Fang, Haoquan and Li, Boyang and Huang, Yushan and Wang, Elvis and Eftekhar, Ainaz and Lee, Jason and Yuan, Wentao and others},
  journal={arXiv preprint arXiv:2505.09990},
  year={2025}
}

@inproceedings{li2024mvbench,
  title={Mvbench: A comprehensive multi-modal video understanding benchmark},
  author={Li, Kunchang and Wang, Yali and He, Yinan and Li, Yizhuo and Wang, Yi and Liu, Yi and Wang, Zun and Xu, Jilan and Chen, Guo and Luo, Ping and others},
  booktitle={Proceedings of the IEEE/CVF Conference on Computer Vision and Pattern Recognition},
  pages={22195--22206},
  year={2024}
}

@inproceedings{song2015sun,
  title={Sun rgb-d: A rgb-d scene understanding benchmark suite},
  author={Song, Shuran and Lichtenberg, Samuel P and Xiao, Jianxiong},
  booktitle={Proceedings of the IEEE conference on computer vision and pattern recognition},
  pages={567--576},
  year={2015}
}

@article{liu2024ocrbench,
  title={Ocrbench: on the hidden mystery of ocr in large multimodal models},
  author={Liu, Yuliang and Li, Zhang and Huang, Mingxin and Yang, Biao and Yu, Wenwen and Li, Chunyuan and Yin, Xu-Cheng and Liu, Cheng-Lin and Jin, Lianwen and Bai, Xiang},
  journal={Science China Information Sciences},
  volume={67},
  number={12},
  pages={220102},
  year={2024},
  publisher={Springer}
}

@article{wang2024charxiv,
  title={Charxiv: Charting gaps in realistic chart understanding in multimodal llms},
  author={Wang, Zirui and Xia, Mengzhou and He, Luxi and Chen, Howard and Liu, Yitao and Zhu, Richard and Liang, Kaiqu and Wu, Xindi and Liu, Haotian and Malladi, Sadhika and others},
  journal={Advances in Neural Information Processing Systems},
  volume={37},
  pages={113569--113697},
  year={2024}
}

@inproceedings{yue2024mmmu,
  title={Mmmu: A massive multi-discipline multimodal understanding and reasoning benchmark for expert agi},
  author={Yue, Xiang and Ni, Yuansheng and Zhang, Kai and Zheng, Tianyu and Liu, Ruoqi and Zhang, Ge and Stevens, Samuel and Jiang, Dongfu and Ren, Weiming and Sun, Yuxuan and others},
  booktitle={Proceedings of the IEEE/CVF conference on computer vision and pattern recognition},
  pages={9556--9567},
  year={2024}
}

@inproceedings{yue2025mmmu,
  title={Mmmu-pro: A more robust multi-discipline multimodal understanding benchmark},
  author={Yue, Xiang and Zheng, Tianyu and Ni, Yuansheng and Wang, Yubo and Zhang, Kai and Tong, Shengbang and Sun, Yuxuan and Yu, Botao and Zhang, Ge and Sun, Huan and others},
  booktitle={Proceedings of the 63rd Annual Meeting of the Association for Computational Linguistics (Volume 1: Long Papers)},
  pages={15134--15186},
  year={2025}
}

@article{lu2023mathvista,
  title={Mathvista: Evaluating mathematical reasoning of foundation models in visual contexts},
  author={Lu, Pan and Bansal, Hritik and Xia, Tony and Liu, Jiacheng and Li, Chunyuan and Hajishirzi, Hannaneh and Cheng, Hao and Chang, Kai-Wei and Galley, Michel and Gao, Jianfeng},
  journal={arXiv preprint arXiv:2310.02255},
  year={2023}
}

@article{xiao2024logicvista,
  title={Logicvista: Multimodal llm logical reasoning benchmark in visual contexts},
  author={Xiao, Yijia and Sun, Edward and Liu, Tianyu and Wang, Wei},
  journal={arXiv preprint arXiv:2407.04973},
  year={2024}
}

@article{jiang2024mmsearch,
  title={Mmsearch: Benchmarking the potential of large models as multi-modal search engines},
  author={Jiang, Dongzhi and Zhang, Renrui and Guo, Ziyu and Wu, Yanmin and Lei, Jiayi and Qiu, Pengshuo and Lu, Pan and Chen, Zehui and Fu, Chaoyou and Song, Guanglu and others},
  journal={arXiv preprint arXiv:2409.12959},
  year={2024}
}

@misc{openclaw2026repo,
  author       = {{OpenClaw}},
  title        = {OpenClaw},
  year         = {2026},
  howpublished = {\url{https://github.com/openclaw/openclaw}},
  note         = {GitHub repository, accessed 2026-04-15}
}

@inproceedings{si2025design2code,
  title={Design2code: Benchmarking multimodal code generation for automated front-end engineering},
  author={Si, Chenglei and Zhang, Yanzhe and Li, Ryan and Yang, Zhengyuan and Liu, Ruibo and Yang, Diyi},
  booktitle={Proceedings of the 2025 Conference of the Nations of the Americas Chapter of the Association for Computational Linguistics: Human Language Technologies (Volume 1: Long Papers)},
  pages={3956--3974},
  year={2025}
}

@article{ge2025advancing,
  title={Advancing vision-language models in front-end development via data synthesis},
  author={Ge, Tong and Liu, Yashu and Ye, Jieping and Li, Tianyi and Wang, Chao},
  journal={arXiv preprint arXiv:2503.01619},
  year={2025}
}

@article{he2026vision2web,
  title={Vision2Web: A Hierarchical Benchmark for Visual Website Development with Agent Verification},
  author={He, Zehai and Hong, Wenyi and Yang, Zhen and Pan, Ziyang and Liu, Mingdao and Gu, Xiaotao and Tang, Jie},
  journal={arXiv preprint arXiv:2603.26648},
  year={2026}
}

@article{xie2025osworld,
  title={Osworld: Benchmarking multimodal agents for open-ended tasks in real computer environments},
  author={Xie, Tianbao and Zhang, Danyang and Chen, Jixuan and Li, Xiaochuan and Zhao, Siheng and Cao, Ruisheng and Toh, Jing Hua and Cheng, Zhoujun and Shin, Dongchan and Lei, Fangyu and others},
  journal={Advances in Neural Information Processing Systems},
  volume={37},
  pages={52040--52094},
  year={2025}
}

@article{rawles2024androidworld,
  title={AndroidWorld: A dynamic benchmarking environment for autonomous agents},
  author={Rawles, Christopher and Clinckemaillie, Sarah and Chang, Yifan and Waltz, Jonathan and Lau, Gabrielle and Fair, Marybeth and Li, Alice and Bishop, William and Li, Wei and Campbell-Ajala, Folawiyo and others},
  journal={arXiv:2405.14573},
  year={2024}
}

@article{he2024webvoyager,
  title={WebVoyager: Building an End-to-End Web Agent with Large Multimodal Models},
  author={He, Hongliang and Yao, Wenlin and Ma, Kaixin and Yu, Wenhao and Dai, Yong and Zhang, Hongming and Lan, Zhenzhong and Yu, Dong},
  journal={arXiv preprint arXiv:2401.13919},
  year={2024}
}

@misc{PinchBench,
  title = {PinchBench},
  howpublished = {\url{https://github.com/pinchbench/skill}},
}

@article{ClawEval,
  title={Claw-Eval: Toward Trustworthy Evaluation of Autonomous Agents},
  author={Ye, Bowen and Li, Rang and Yang, Qibin and Liu, Yuanxin and Yao, Linli and Lv, Hanglong and Xie, Zhihui and An, Chenxin and Li, Lei and Kong, Lingpeng and others},
  journal={arXiv preprint arXiv:2604.06132},
  year={2026}
}

@misc{ZClawBench,
  title = {ZClawBench},
  howpublished = {\url{https://huggingface.co/datasets/zai-org/ZClawBench}},
}

@article{MMSearch,
  title={Mmsearch: Benchmarking the potential of large models as multi-modal search engines},
  author={Jiang, Dongzhi and Zhang, Renrui and Guo, Ziyu and Wu, Yanmin and Lei, Jiayi and Qiu, Pengshuo and Lu, Pan and Chen, Zehui and Fu, Chaoyou and Song, Guanglu and others},
  journal={arXiv preprint arXiv:2409.12959},
  year={2024}
}

@article{MMSearch-Plus,
  title={Mmsearch-plus: Benchmarking provenance-aware search for multimodal browsing agents},
  author={Tao, Xijia and Teng, Yihua and Su, Xinxing and Fu, Xinyu and Wu, Jihao and Tao, Chaofan and Liu, Ziru and Bai, Haoli and Liu, Rui and Kong, Lingpeng},
  journal={arXiv preprint arXiv:2508.21475},
  year={2025}
}

@inproceedings{SimpleVQA,
  title={Simplevqa: Multimodal factuality evaluation for multimodal large language models},
  author={Cheng, Xianfu and Zhang, Wei and Zhang, Shiwei and Yang, Jian and Guan, Xiangyuan and Wu, Xianjie and Li, Xiang and Zhang, Ge and Liu, Jiaheng and Mai, Yuying and others},
  booktitle={Proceedings of the IEEE/CVF International Conference on Computer Vision},
  pages={4637--4646},
  year={2025}
}

@misc{BrowsecompVL,
      title={WebWatcher: Breaking New Frontier of Vision-Language Deep Research Agent}, 
      author={Xinyu Geng and Peng Xia and Zhen Zhang and Xinyu Wang and Qiuchen Wang and Ruixue Ding and Chenxi Wang and Jialong Wu and Yida Zhao and Kuan Li and Yong Jiang and Pengjun Xie and Fei Huang and Jingren Zhou},
      year={2025},
      eprint={2508.05748},
      archivePrefix={arXiv},
      primaryClass={cs.IR},
      url={https://arxiv.org/abs/2508.05748}, 
}

@misc{FACTS,
    author = {Alon Jacovi and Andrew Wang and Chris Alberti and Connie Tao,  Jon Lipovetz and Kate Olszewska and Lukas Haas and Michelle Liu and Nate Keating and Adam Bloniarz and Carl Saroufim and Corey Fry and Dror Marcus and  Doron Kukliansky and Gaurav Singh Tomar and James Swirhun and Jinwei Xing and Lily Wang and Michael Aaron and Moran Ambar and Rachana Fellinger and Rui Wang and Ryan Sims and Zizhao Zhang and Sasha Goldshtein and Yossi Matias and Dipanjan Das},
    title = {FACTS Leaderboard},
    year = {2024},
    howpublished = {\url{https://kaggle.com/facts-leaderboard}},
    note = {Google DeepMind, Google Research, Google Cloud, Kaggle}
}

@misc{wu2023vguidedvisualsearch,
      title={V*: Guided Visual Search as a Core Mechanism in Multimodal LLMs}, 
      author={Penghao Wu and Saining Xie},
      year={2023},
      eprint={2312.14135},
      archivePrefix={arXiv},
      primaryClass={cs.CV},
      url={https://arxiv.org/abs/2312.14135}, 
}

@inproceedings{henry2020query,
  title={Query-key normalization for transformers},
  author={Henry, Alex and Dachapally, Prudhvi Raj and Pawar, Shubham Shantaram and Chen, Yuxuan},
  booktitle={Findings of the Association for Computational Linguistics: EMNLP 2020},
  pages={4246--4253},
  year={2020}
}

@article{yang2025ui2code,
  title={UI2Code\^{} N: UI-to-Code Generation as Interactive Visual Optimization},
  author={Yang, Zhen and Hong, Wenyi and Xu, Mingde and Fan, Xinyue and Wang, Weihan and Cheng, Jiele and Gu, Xiaotao and Tang, Jie},
  journal={arXiv preprint arXiv:2511.08195},
  year={2025}
}
